\documentclass[sigconf]{acmart} 

\usepackage{amsmath}
\usepackage{algorithmic, algorithm} 

\usepackage{booktabs}       
\usepackage{threeparttable}
\usepackage{multirow}       
\usepackage[caption=false,font=normalsize,labelfont=sf,textfont=sf]{subfig} 

\usepackage{graphicx}
\usepackage{float}
\usepackage{url}
\usepackage{verbatim}
\usepackage{enumitem}
\usepackage{wrapfig}

\usepackage{pifont}

\AtBeginDocument{%
  }

\setcopyright{none}
\copyrightyear{}
\acmYear{}
\acmDOI{}
\acmConference[Conference acronym 'XX]{Make sure to enter the correct
  conference title from your rights confirmation email}{June 03--05,
  2018}{Woodstock, NY}
\acmISBN{978-1-4503-XXXX-X/2018/06}
\renewcommand\footnotetextcopyrightpermission[1]{}

\begin{document}

\title{BamaER: A Behavior-Aware Memory-Augmented Model for Exercise Recommendation}

\author{Qing Yang}
\orcid{0000-0003-2844-5098}
\affiliation{
  \institution{Guilin University of Electronic Technology}
  \city{Guilin}
  \country{China}
}
\email{yangqing@guet.edu.cn}

\author{Yuhao Jiang}
\orcid{0009-0007-4163-6143}
\affiliation{
  \institution{Guilin University of Electronic Technology}
  \city{Guilin}
  \country{China}
}
\email{jiangyu@mails.guet.edu.cn}

\author{Rui Wang}
\affiliation{
  \institution{Jinan University}
  \city{Guangzhou}
  \country{China}
}
\email{wangrui377@outlook.com}

\author{Jipeng Guo}
\orcid{0009-0007-4880-0305}
\affiliation{
  \institution{Beijing University}
  \city{Beijing}
  \country{China}
}
\email{guojipeng@buct.edu.cn}

\author{Yejiang Wang}
\orcid{0000-0001-7908-4275}
\affiliation{
  \institution{Xiamen university}
  \city{Xiamen}
  \country{China}
}
\email{wyejiang@gmail.com}

\author{Xinghe Cheng}
\orcid{0000-0001-9432-5794}
\affiliation{
  \institution{Jinan University}
  \city{Guangzhou}
  \country{China}
}
\email{jnuchengxh@hotmail.com}

\author{Zezheng Wu}
\orcid{0009-0001-9351-1314}
\affiliation{
  \institution{Guilin University of Electronic Technology}
  \city{Guilin}
  \country{China}
}
\email{zezhengwu@hotmail.com}

\author{Jiapu Wang}
\orcid{0000-0001-7639-5289}
\affiliation{
  \institution{Nanjing University of Science and Technology}
  \city{Nanjing}
  \country{China}
}
\email{jiapu.wang@njust.edu.cn}

\author{Jingwei Zhang}
\orcid{0000-0002-7907-2853}
\affiliation{
  \institution{Guilin University of Electronic Technology}
  \city{Guilin}
  \country{China}
}
\email{gtzjw@hotmail.com}

\renewcommand{\shortauthors}{Trovato et al.}

\begin{abstract}
    Exercise recommendation focuses on personalized exercise selection conditioned on students’ learning history, personal interests, and other individualized characteristics.
    Despite notable progress, most existing methods represent student learning solely as exercise sequences, overlooking rich behavioral interaction information. This limited representation often leads to biased and unreliable estimates of learning progress. Moreover, fixed-length sequence segmentation limits the incorporation of early learning experiences, thereby hindering the modeling of long-term dependencies and the accurate estimation of knowledge mastery. 
    To address these limitations, we propose BamaER, a \underline{\textbf{B}}ehavior-\underline{\textbf{a}}ware \underline{\textbf{m}}emory-\underline{\textbf{a}}ugmented \underline{\textbf{E}}xercise \underline{\textbf{R}}ecommendation framework that comprises three core modules: (i) the learning progress prediction module that captures heterogeneous student interaction behaviors via a tri-directional hybrid encoding scheme; (ii) the memory-augmented knowledge tracing module that maintains a dynamic memory matrix to jointly model historical and current knowledge states for robust mastery estimation; and (iii) the exercise filtering module that formulates candidate selection as a diversity-aware optimization problem, solved via the Hippopotamus Optimization Algorithm to reduce redundancy and improve recommendation coverage.
    Experiments on five real-world educational datasets show that BamaER consistently outperforms state-of-the-art baselines across a range of evaluation metrics.
\end{abstract}




\maketitle

\section{Introduction}
With the development of online intelligent education systems and advances in information technology, students now have convenient access to a large volume of learning resources~\cite{HuaLZ2019, SheYL2024}. However, the available exercise resources vary widely in knowledge coverage, difficulty level, and instructional design~\cite{GuaXC2023, HuoWN2020, WanKL2024}. At the same time, students often lack an accurate assessment of their own knowledge mastery, making it difficult to select exercises that match their current learning needs. Therefore, accurately recommending appropriate exercises to students has become a key challenge in intelligent education.
\begin{figure}[t]
    \centering
    \includegraphics[width=0.48\textwidth]{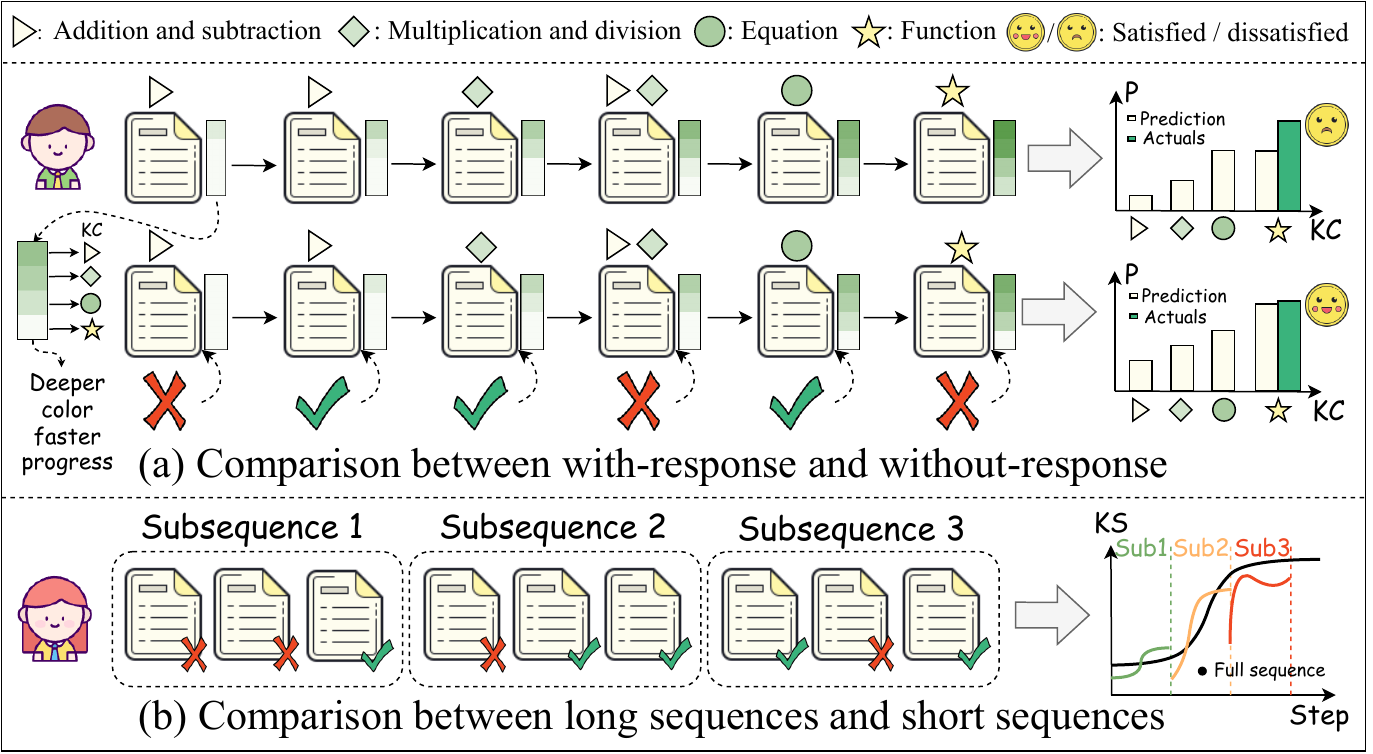}
    \caption{Illustration of key challenges.  (a) Learning progress depends on response accuracy: perfect scorers advance faster than those who make mistakes. Excluding this, models may equate all students’ progress. (b) Existing methods’ non-interactive fixed-length subsequence splits lose long-range dependencies, limiting short-term context models’ accuracy in estimating students’ current knowledge.}
    \label{fig:Examples of two problems}
\end{figure}

\looseness=-1
Existing methods typically estimate the probability of each knowledge concept being practiced next based solely on exercise sequences, and use these predicted probabilities as proxies for learning progress~\cite{RenLS2023, WuLT2020}. However, such approaches often overlook the rich interaction patterns embedded in the learning process, limiting their ability to capture students’ behavioral dynamics. As illustrated in Figure~\ref{fig:Examples of two problems} (a), when students’ learning interaction behaviors are not taken into account, learning progress can only be assessed based on the number of responses to different knowledge concepts, the models that rely solely on exercise order fail to distinguish students with different response patterns, treating them as having similar mastery levels. This mismatch leads to a \textit{learning progress illusion}, undermining the accurate modeling of student learning trajectories.

\looseness=-1
Beyond this limitation, a further challenge arises in preserving the continuity of students’ learning trajectories while estimating their mastery levels at each exercise.
A widely adopted approach first infers a student’s mastery of underlying knowledge concepts, and then predicts performance on corresponding exercises~\cite{LiuRG2025, RenLS2023, WuLT2020}.
To standardize input and improve training efficiency, most methods segment a student's exercise history into fixed-length subsequences~\cite{GuaXC2023, PieBH2015,WanWG2024}, with remaining interactions forming a separate segment.
Although effective computationally, this strategy fragments students’ interaction histories~\cite{HeLW2023}, scattering concept-related signals across disjoint sequences. This fragmentation undermines the model’s capacity to accurately track mastery progression and compromises the reliability of personalized recommendations.
As illustrated in Figure~\ref{fig:Examples of two problems} (b), temporally distant interactions on the same concept may be split into different subsequences, disrupting concept-level learning continuity.

\looseness=-1
In particular, the inability to model long-range dependencies across temporally distant interactions further exacerbates this issue.
Capturing the long-range impact of early learning experiences remains a critical challenge. RNN-based~\cite{WuLT2020} and attention-based~\cite{RenLS2023} models typically rely on the continuity of input sequences to encode temporal dependencies.
However, fixed-length segmentation often truncates crucial historical interactions, hindering the model’s ability to incorporate early learning behaviors into current mastery estimation~\cite{HeLW2023, WanCW2024}.
This segmentation-induced discontinuity discards valuable contextual signals and ultimately degrades the accuracy of knowledge state prediction.

\begin{figure*}[h]
    \centering
    \includegraphics[width=1\textwidth]{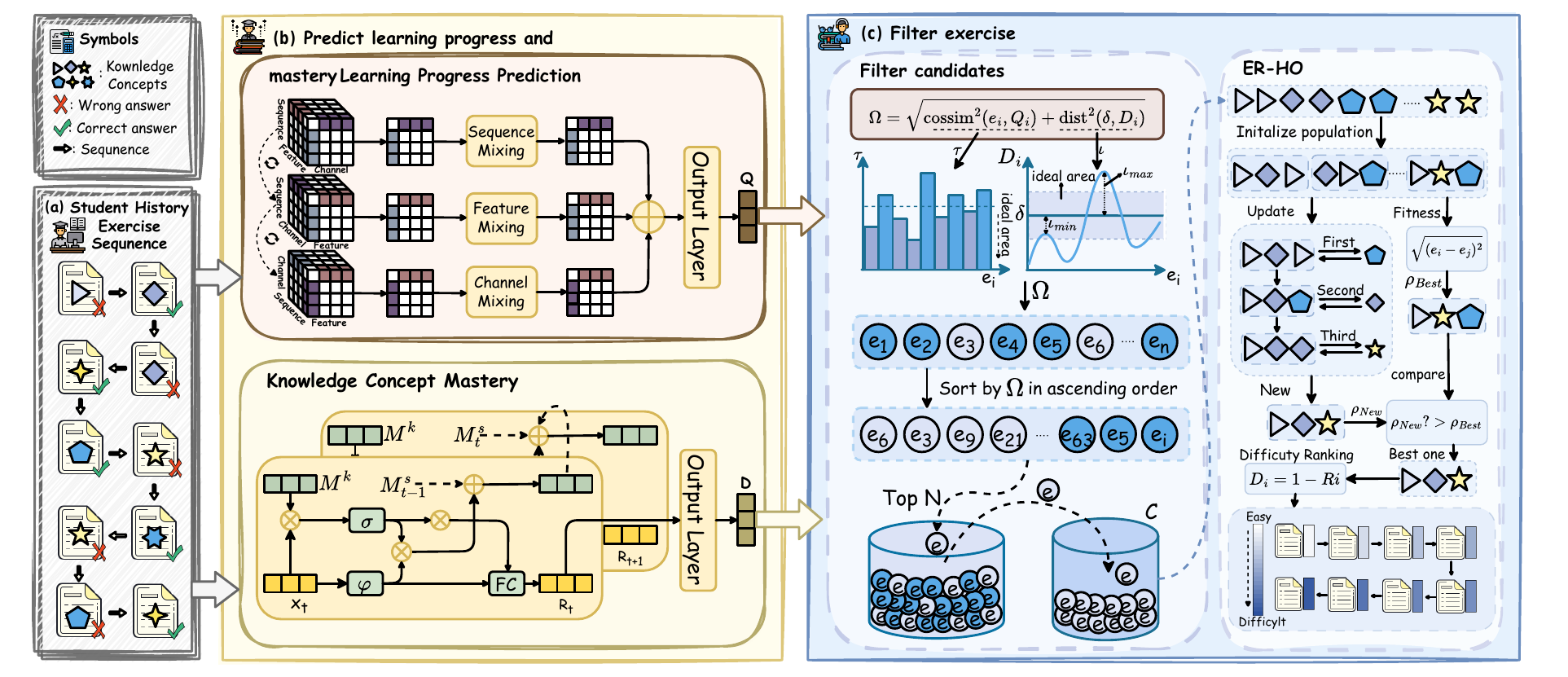}
    \caption{Overall architecture of BamaER. (a) We take the student’s historical exercise sequence as input, which is arranged in the order of interactions and includes the knowledge concept of each exercise together with the corresponding response outcome. (b) We then input the  sequence into the learning progress prediction module to and the knowledge concept mastery prediction module. These two modules embed features such as exercises, knowledge concepts, and scores into the model, thereby generating the learning progress and the knowledge concept mastery for each student. (c) These two matrices are input into the exercise filter module, where they serve as criteria for selecting a candidate set from the exercise repository. The candidate set is then optimized using the ER-HO algorithm, ultimately generating the exercise recommendation list.}
    
    \label{fig:BamaER}
\end{figure*}

\looseness=-1
To address the above challenges, in this paper, we propose a novel \underline{\textbf{B}}ehavior-\underline{\textbf{a}}ware \underline{\textbf{m}}emory-\underline{\textbf{a}}ugmented \underline{\textbf{E}}xercise \underline{\textbf{R}}ecommendation (BamaER) method. Specifically, we first design a learning progress prediction module that integrates complex behavioral interactions of students to help improve the accuracy of predicting learning progress.
Subsequently, we develop a knowledge concept mastery prediction module that combines a memory module and a weighting module to respectively capture long-term learning accumulation and recent learning dynamics of students, enabling precise prediction of mastery levels.
Meanwhile, we develop an exercise filter module that utilizes the predictive results from the first two modules to select exercises.
Moreover, we introduce hippopotamus optimization algorithm to further filter the exercises, thereby enhancing the diversity of the exercise list. 

The main contributions of this paper are summarized as follows:

\begin{itemize}[left=0cm]
    \item We proposes a \underline{\textbf{B}}ehavior-\underline{\textbf{a}}ware \underline{\textbf{m}}emory-\underline{\textbf{a}}ugmented \underline{\textbf{E}}xercise \underline{\textbf{R}}ecommendation (BamaER) method that integrates multiple modules to jointly predict students’ learning progress and knowledge mastery, while leveraging optimization strategies to deliver personalized exercise recommendations;
    \item We design a learning progress prediction module which embeds complex interaction information of students into the input and employs a tri-directional hybrid input approach to enhance information extraction capability;
    \item We propose a memory-augmented knowledge tracing module that maintains a dynamic memory matrix to track students’ evolving knowledge states and accurately predict their mastery of future exercises.
\end{itemize}

\looseness=-1

\section{Related Work}
In this section, we review the related work from the literature on exercise recommendation and knowledge tracing.
\subsection{Exercise Recommendation}

In the field of education, exercise recommendation has received significant attention~\cite{CheZF2025}.
Traditional recommendation approaches can mainly be categorized into content-based filtering (CBF)~\cite{PazB2007,WalRL2004}, collaborative filtering (CF)~\cite{SarKK2001,WanDR2006,FanWZ2024}, and hybrid filtering (HF)~\cite{ChaLC2016,KlaIN2015,GeoL2019}.
Most of these methods utilize similarity measures to provide exercise students.
However, these methods lack the ability to perceive knowledge state of students, which limits the accuracy of the recommended exercises.
With the development of reinforcement learning, Huang \textit{et al.}~\cite{HuaLZ2019} employed reinforcement learning techniques to recommend exercises for students by simultaneously optimizing three objectives.
Liu \textit{et al.}~\cite{LiuHL2023} designed a multi-agent exercise recommendation module based on reinforcement learning, where knowledge concepts act as intelligent agents engaging in mutual competition.
In addition, the KG4EX~\cite{GuaXC2023} utilizes knowledge graphs to offer comprehensive explanations for exercise recommendations.
However, these methods primarily optimize for accuracy, while neglecting the comprehensive consideration and evaluation of other crucial objectives such as diversity and novelty. 
Wu \textit{et al.}~\cite{WuLT2020} proposed the KCP-ER model, which divides the recommendation process into into a prediction phase and a filtering phase, thereby selecting exercises with appropriate difficulty levels for students.
The two-layer exercise recommendation model MulOER-SAN~\cite{RenLS2023} employs a self-attention mechanism to enhance performance.
However, both methods fail to effectively utilize the complex behavioral interaction information in the exercise sequences when predicting knowledge state of students.
As a result, they struggle to capture behavioral patterns of students accurately, leading to less accurate knowledge state predictions and consequently casing suboptimal recommendations.

\subsection{Knowledge Tracing}
Knowledge Tracing (KT) is a core technology in intelligent education systems~\cite{LiDQ2025}, whose core goal is to real-time track and evaluate students' mastery of specific knowledge points~\cite{CheGX2025,SonLC202}, providing key support for exercise recommendation~\cite{LiuWB2022,AbdWN2023}.
Specifically, the mainstream methods of KT can be divided into three categories: First, methods based on Recurrent Neural Networks (RNNs), represented by Deep Knowledge Tracing (DKT)~\cite{PieBH2015}. These methods adopt Long Short-Term Memory (LSTM) networks~\cite{GraG2012} to process learning sequences but struggle to efficiently capture long-term dependencies.
Secondly, Memory-Augmented Neural Networks (MANNs), which optimize the modeling of long-term dependencies by introducing single or dual memory matrices~\cite{SanBB2016, SukWF2015}. Dynamic Key-Value Memory Network (DKVMN)~\cite{ZhaSK2017} is a typical example, as it encodes knowledge points using a key memory matrix and dynamically updates students' knowledge states.
Thirdly, Attention Neural Networks (ANNs), developed based on the attention mechanism~\cite{VasSP2017}. Models such as SAKT~\cite{PanK2019} and AKT~\cite{GhoHL2020} leverage self-attention to capture long-term dependencies in learning behaviors, with performance superior to the previous two categories. However, due to the weight mechanism’s tendency to emphasize recently learned knowledge~\cite{HeLW2023}, they still have certain limitations in specific exercise recommendation scenarios.

\section{PROBLEM AND FRAMEWORK OVERVIEW}
This section commences with a formal definition of the exercise recommendation problem and proceeds to outline the architecture of the BamaER framework.

\subsection{Problem Statement}  
In an intelligent education system, suppose there are $N_{s}$ students and an exercise bank EB. 
For a student about to start a new exercise, the historical learning records are denoted as $H=\{(\boldsymbol{e}_{1}, r_{1}),(\boldsymbol{e}_{2}, r_{2}),\\ \cdots,(\boldsymbol{e}_{i}, r_{i})\}$, where $\boldsymbol{e}_{i}$ represents the $i$-th exercise of student, and $r_{i}$ denotes the corresponding performance.
Generally, if the student answers exercise $\boldsymbol{e}_{i}$ correctly, $r_{i}$ equals to 1, otherwise $r_{i}$ equals to 0.
The contained knowledge concepts are represented as $K=\{k_{1},k_{2},\ldots,k_{N_{k}}\}$.  
In addition, we represent each exercise as $\boldsymbol{e}_{i}=[\varphi_{1},\varphi_{2},\cdots,\varphi_{{N_{k}}}]$, if the exercise $e_i$ involves the $j$-th knowledge concept, $\varphi_{j}$ equals to 1, otherwise $\varphi_{j}$ equals to 0.
Then, the problem we study in this paper is described as follows: Given exercise record $H$ of a student and the corresponding EB, our objective is to recommend a personalized exercise recommendation list $L$ for the student.
For ease of illustration, Table \ref{tab:notations} summarizes the key symbols and their corresponding descriptions.

\begin{table}[t]
    \centering
    \caption{Key symbols and descriptions.}
    \label{tab:notations}
    \begin{tabular}{c l}
        \toprule
        \textbf{Notations} & \textbf{Description} \\ 
        \midrule 
        EB & Set of exercises, $N_{e}$ is the number of exercises. \\ 
        $H$ & Exercise log of the student. \\
        $K$ & All knowledge concepts contained in the exercises. \\
        $\boldsymbol{e}_{i}$ & i-th exercise. \\ 
        $r_{i}$ & Score of the student on the $i$-th exercise. \\ 
        $Q$ & Knowledge concept coverage vector. \\ 
        $Z$ & Knowledge concept mastery vector. \\ 
        $R$ & Mastery level of students for the exercises. \\ 
        $D$ & Difficulty of exercises. \\ 
        $L$ & Exercise recommendation list. \\ 
        \bottomrule
    \end{tabular}
\end{table}

\subsection{Framework Overview}
Figure \ref{fig:BamaER} illustrates the overall framework of BamaER, which is primarily composed of three modules: the learning progress prediction module, the knowledge concept mastery prediction module, and the exercise filter module.
The model takes historical response records of students as input.
Firstly, we design a learning progress prediction module to assess the learning progress of students.
Next, in the knowledge concept mastery prediction module, we design a knowledge tracing model to estimate the mastery of students for various knowledge concepts.
Finally, in the exercise filter module, the learning progress and knowledge concept mastery extracted from the previous two modules are used to filter and generate a candidate exercise set from EB.
The sine hippopotamus algorithm within this module is applied to further filter the candidate exercise set, optimizing the diversity of the recommendation list, and ultimately producing the final exercise recommendation list $L$.

\section{Methodology}
In this section, we detail the structure of BamaER, focusing on its three key modules.
Firstly, to more accurately predict learning progress of students, we separately design a learning progress prediction module.
Secondly, to more accurately predict knowledge concept mastery level of students, we design a knowledge concept mastery level prediction module.
Finally, we design an exercise filter module to select suitable exercises, and incorporate the hippopotamus optimization algorithm into this module to enhance the diversity of the recommendation list.

\subsection{Learning Progress Prediction}
To predict learning progress of students, we construct learning progress prediction module.
It is typically measured by the coverage rate of knowledge concepts related to the exercises completed by the student~\cite{RenLS2023}.
Therefore, based on the knowledge concepts sequence from the first step to the $i$-th step, we infer the probability of each concept being encountered in the subsequent step.
The specific process is as follows:

\looseness=-1
\textbf{Embedding Layer}.
In the learning progress prediction module, to construct the representations of knowledge concepts and student interaction information, we first map the discrete knowledge concept into dense vector $\boldsymbol{k} \in \mathbb{R}^{N_{k}\times d}$, where $N_{k}$ indicates the total knowledge concepts, $d$ denotes the embedding size, and we map the student interaction data (e.g., response outcomes) into another dense vector $\boldsymbol{r} \in \mathbb{R}^{n\times d}$, with $n$ representing the possible interaction values~\cite{CheKH2016}. 
Then we concatenate them along the specified dimension into a single embedding matrix, where embedding vectors are organized row-wise, and the columns denote channels.
By further stacking all embedding matrix together, we obtain a three-dimensional embedding matrix $X\in\mathbb{R}^{m\times l\times v}$~\cite{GaoZL2024}, where $m$  corresponds to the channel dimension, $l$ denotes the sequence length, and $v$ denotes the number of interaction features, which we refer to as the KC-interaction matrix , as illustrated in the corresponding part of the module in Figure~\ref{fig:BamaER}.
Since performing normalization during the subsequent fusion process would slow down training, we apply normalization before information extraction:
\begin{equation}
X'=\text{LayerNorm}(X),
\end{equation}
where $\text{LayerNorm}(\cdot)$ denotes the Layer Normalization~\cite{BaKH2016}.

\looseness=-1
\textbf{Information Extraction}.
After constructing the KC-interaction matrix after layer normalization $X'$, we need to extract key information from it.
The first step is to capture the positional information of each knowledge concept in the sequence, in order to model the sequential dependencies within the knowledge concept sequence.
We use the rows along the sequence direction of $X'$ as input $P=[\boldsymbol{p}_1,\boldsymbol{p}_2,\dots,\boldsymbol{p}_l]$ and outputs vectors with the same dimension. The output of the $n$-th layer is represented as $\boldsymbol{y_p}$:
\begin{equation} 
\boldsymbol{y}_{p} = W_{2} \cdot g\big( W_{1} \cdot P + \boldsymbol{b}_{1} \big) + \boldsymbol{b}_{2},
\end{equation}
where $W_{1}$ and $W_{2}$ represent the learnable weights of the different fully connected layers, $\boldsymbol{b}_i$ denote the corresponding bias vectors, $r$ is the hidden size.
$g(\cdot)$ denotes the non-linear activation function GELU~\cite{HenG2016}.

\looseness=-1
Secondly, to more effectively integrate student interaction information into the representation of each position, it is necessary to model the relationships among interaction features.
To achieve this, we take the columns along the feature dimension of $X'$ as input $F=[\boldsymbol{f}_1,\boldsymbol{f}_2,\dots,\boldsymbol{f}_v]$.
The result produced by the $n$-th layer within the feature mixing branch is denoted by:
\begin{equation} 
\boldsymbol{y}_{f} = W_{4} \cdot g\big( W_{3} \cdot F + \boldsymbol{b}_{3} \big) + \boldsymbol{b}_{4},
\end{equation}
where $W_{3}$ and $W_{4}$ represent the learnable weights of the different fully connected layers.

\looseness=-1
In addition, the embeddings of knowledge concepts or interaction information contain different latent semantics across dimensions. Therefore, learning the internal correlations among the dimensions of these embedding vectors is also essential. To this end, we take the columns along the channel dimension of $X'$ as input $C=[\boldsymbol{c}_1,\boldsymbol{c}_2,\dots,\boldsymbol{c}_m]$.
The output of the $n$-th layer of this module can be represented as:
\begin{equation} 
\boldsymbol{y}_{c} = W_{6} \cdot g\big( W_{5} \cdot C + \boldsymbol{b}_{5} \big) + \boldsymbol{b}_{6},
\end{equation}
where $\boldsymbol{y}_{c}$ represents the output, $W_{5}$ and $W_{6}$ denote the learnable weights of the different fully connected layers.

\looseness=-1
Finally, to ensure that the model captures cross-dimensional information for recommendation, we use a simple summation to aggregate the three types of information as the output of this layer~\cite{GaoZL2024}.
The information fusion operation is formulated as:
\begin{equation} 
Y = \boldsymbol{y}_{p}^\top \oplus \boldsymbol{y}_{f}^\top \oplus \boldsymbol{y}_{c}^\top,
\end{equation}
where $\oplus$ denotes vector addition, and $\top$ denotes transposition.

\looseness=-1
\textbf{Training}.
When predicting the next knowledge concept, we extract the hidden state associated with the target knowledge concept and compute its dot product with the embeddings of all candidate knowledge concepts to measure similarity.
We employ cross-entropy loss as the loss function for this module.
\begin{eqnarray} 
    \begin{split}
        \mathcal{L} = -\sum_{S_{n} \in S} \sum_{t \in [1, \ldots, T]} \Biggr[ 
            \log \left(\sigma\left(r_{i, t}\right)\right)+ \sum_{j \notin S_{n}} \log \left(1 - \sigma\left(r_{j, t}\right)\right) \Biggr],\nonumber
    \end{split}
\end{eqnarray}
where $\sigma(\cdot)$ denotes the sigmoid function~\cite{ElfUD2018}, $r_{i, t}$ is the predicted similarity of model to ground-truth knowledge concept $i_{t}$, and $r_{j, t}$ is the predicted similarity to negative sampled knowledge concepts at position $t$.

\looseness=-1
\textbf{Prediction}.
Finally, the model produces a vector whose length equal to the total number of knowledge concepts, denoted by: $\boldsymbol{o}=[o_{1},\cdots, o_{i},\cdots,o_{N_k}]$, where each of its elements is the probability of occurrence of the corresponding knowledge concept.

\looseness=-1
Since the knowledge concepts included in future exercises should primarily be those that were previously answered incorrectly to reinforce mastery, our objective is to reduce the frequency of concepts that have already been correctly answered in subsequent exercises.
To achieve this, we integrate a vector of weights with the module’s output.
This vector spans the knowledge concept space and is formally defined as:
$\boldsymbol{w}=[w_{1}, w_{2}, \cdots, w_{i}]$, where $w_{i}$ denotes the weight of each knowledge concept, and its formula can be expressed as:
\begin{eqnarray} 
    w_{i} = \begin{cases} 
      1 - \frac{\alpha _i}{\nu _i}, & \text{if } \nu _i > 0, \\
      1, & \text{if } v_i = 0,
   \end{cases}
\end{eqnarray}
which $\alpha _i$ denotes the number of correct responses for knowledge concept $c_i$, and $\nu _i$ represents its occurrence count.
Then, by combining the output of the module with the weighting vector, the learning progress of student related to the knowledge concepts can be obtained: 
\begin{eqnarray} 
    q_i=o_{i}w_{i},
\end{eqnarray}
where $q_i$ is the $i$-th element of the learning progress $Q$.

\subsection{Knowledge Concept Mastery Prediction}
To predict the mastery level of each knowledge concept, we design a knowledge tracing model.
This model consists of memory update and context knowledge modeling.
The first part employs a memory matrix to capture and preserve the previous knowledge state.
The second part employs a self-attention mechanism for context-aware processing, enabling it to represent the current knowledge state.
The detailed steps are as follows:

\looseness=-1
\textbf{Embedding Layer}. We take historical interaction sequence of students as inputs.
Then, we construct the exercise embedding $\boldsymbol{u}$ and the interaction embedding $\boldsymbol{g}$.
Specifically, $\boldsymbol{u}$ participates in the prediction process together with the current knowledge state of student, while $\boldsymbol{g}$ is used to update the present knowledge state of student.
The representations of these two embeddings are as follows:
\begin{equation} 
\begin{aligned}
    \boldsymbol{u}&=\text{Softmax}(\boldsymbol{k} \parallel (\lambda \cdot \boldsymbol{e})),\\
    \boldsymbol{g}&=\text{Softmax}({(\boldsymbol{k}\oplus\boldsymbol{r})}\parallel(\lambda \cdot (\boldsymbol{e}\oplus\boldsymbol{r}))),
\end{aligned}
\end{equation}
where $\boldsymbol{k} \in \mathbb{R}^{d}$ represents the embedding of the knowledge concepts, and $\boldsymbol{e} \in \mathbb{R}^{d}$ represents the embedding of exercises, $\boldsymbol{e} \in \mathbb{R}^{d}$ represents the embedding of the response,
$\lambda$ is a learnable parameter.
$\text{Softmax}(\cdot)$ refers to the softmax function~\cite{NwaIG2018}.

\looseness=-1
\textbf{Context Knowledge Modeling}.
Since one exercise state cannot adequately represent the  comprehensive knowledge state of student, incorporating contextual information is crucial for more precise modeling.
Here, we adopt the modeling approach of AKT~\cite{GhoHL2020} to capture contextual knowledge.
A self-attention mechanism is utilized to encode $\boldsymbol{u}$ and generate a context-aware exercise embedding sequence $\hat{\boldsymbol{u}}$.
The specific formulation is as follows:
\begin{eqnarray} 
\boldsymbol{\hat{\boldsymbol{u}}}=\sum_{i=0}^{t}\boldsymbol{v}_{t-i} \cdot a_{t-i},
\end{eqnarray}
where $\boldsymbol{v}_{t-i}$ represents the value embedding, and $a_{t-i}$ represents the attention weight.

\looseness=-1
We apply the same self-attention to encode $\boldsymbol{g}$ and generate a context-aware interaction embedding sequence $\hat{\boldsymbol{g}}$.
The specific formulation is as follows:
\begin{eqnarray} \boldsymbol{\hat{g}}=\sum_{i=0}^{t}\boldsymbol{v}_{t-i} \cdot a_{t-i}.
\end{eqnarray}

\looseness=-1
Then, we retrieve the knowledge state $\boldsymbol{s}_t$ of the current exercise from the $\hat{\boldsymbol{u}}$ and $\hat{\boldsymbol{g}}$:
\begin{equation}
\begin{aligned}
\boldsymbol{\boldsymbol{s}}_t=\sum_{i=1}^{t}\hat{\boldsymbol{v}}_{t-i} \cdot \hat{a}_{t-i},\quad
\hat{a}_{t-i}=\operatorname{Softmax}\left(\frac{\hat{\mathbf{q}}_{t}^{\top} \hat{\mathbf{k}}_{t-i}}{\sqrt{d}}\right),
\end{aligned}
\end{equation}
where $\hat{\boldsymbol{v}}_{i}$ represents the value embedding derived from $\hat{\boldsymbol{g}}$, $\hat{\mathbf{q}}_{t}$ and $ \hat{\mathbf{k}}_{t}$ denote the query embedding and key embedding derived from $\hat{\boldsymbol{u}}$, $d$ denotes the dimension.

\looseness=-1
\textbf{Memory Update}. To retrieve relevant knowledge from the existing memory of student and update it accordingly.
We obtain the key-memory matrix $M^{k}$ and knowledge state matrix of student $M_t^s$ from the memory module. Firstly, we use $M^{k}$ to obtain the the correlation weight vector of the exercise, the specific formula is as follows:
\begin{eqnarray} 
    \boldsymbol{w}_{t}=\text{Softmax}(M^{k}\boldsymbol{u}).
\end{eqnarray}

Subsequently, Subsequently, we update $M_t^s$ through combining $\boldsymbol{w}_{t}$ with the previously stored memory $M_{t-1}^s$:
\begin{eqnarray} 
    M_t^s=\partial(\boldsymbol{s}_{t}\otimes \boldsymbol{w}_{t})\oplus(1-\partial)M_{t-1}^s,
\end{eqnarray}
where $\partial$ denotes the memory updating weight.
Finally, the model retrieves $\boldsymbol{m}_t$, containing the relevant conceptual knowledge linked to the present exercise, the specific formula is as follows:
\begin{eqnarray} 
    \boldsymbol{m}_t=\boldsymbol{w}_{t}M_t^s.
\end{eqnarray}

\looseness=-1
Finally, we correlate the representation vector $\boldsymbol{m}_t$ and the recent knowledge state of student $\boldsymbol{\boldsymbol{u}}_t$ vertically linked, and then fed into a fully connected layer by a Tanh function~\cite{NwaIG2018} to generate a vector $\boldsymbol{\theta}_{t}$.
Subsequently, $\boldsymbol{\theta}_{t}$ is input into another fully connected layer with a sigmoid function~\cite{ElfUD2018} to predict the probability of correctly answering each knowledge concept, the specific formula is as follows:
\begin{equation}
\begin{aligned} 
    \boldsymbol{\theta }_{t}=\text{Tanh}\left(\text{FC}(\boldsymbol{m}_t \parallel \boldsymbol{\boldsymbol{s}}_t)\right),\quad
Z=\text{Sigmoid}\left(\text{FC}(\boldsymbol{\theta }_{t})\right),
\end{aligned}
\end{equation}
where $\text{FC}(\cdot)$ denotes a fully connected layer, $\parallel$ represents the concatenation of two vectors, $Z$ represents the mastery level for all knowledge concept.

\looseness=-1
Since an exercise includes one or more knowledge concepts, we can estimate the likelihood that a student will answer the exercise correctly by considering their mastery level of the relevant knowledge concepts:
\begin{eqnarray} 
    R_{i} = \prod_{i=1}^n \left( z_i \mid k_i = 1 \right),
\end{eqnarray}
where $k_i$ is the $i$-th knowledge concept in exercise $\boldsymbol{e}_{i}$, $z_i$ represents the student's mastery of the i-th knowledge concept, $z_i \in Z$.
Specifically, we represent the proficiency of student with the exercise by the product of their mastery of the knowledge concepts it contains.
This product is then used to quantify the difficulty of the exercise for the student: $D_{i} = 1 - R_{i}$.

\subsection{Exercise Filter}
After predicting students’ learning progress and knowledge concept mastery, we design an exercise filtering module to select suitable exercises from the exercise bank. The module filters via cosine similarity and a distance function: cosine similarity measures the similarity between knowledge concepts in exercise $\boldsymbol{e}{i}$ and the student’s learning progress $Q$, while the distance function calculates the difference between $D{i}$ and expected difficulty $\delta$. The Euclidean norm of these two values is expressed as follows:
\begin{eqnarray} 
    \Omega_{i} = \sqrt{\text{cossim}^{2}(\boldsymbol{e}_{i},Q_{i}) + \text{dist}^{2}(\delta ,D_{i})}.
\end{eqnarray}

\looseness=-1
We arrange the exercises in the exercise bank EB in ascending order according to $\Omega_{i}$, and select the top 200 exercises to form our candidate set.

\looseness=-1
After the preliminary filtering stage, the candidate set may still contain redundant exercises associated with the same knowledge points. To eliminate these redundancies and enhance the diversity of the final recommendation list, this study further incorporates the Hippopotamus Optimization algorithm (HO) to conduct in-depth filtering and optimization of the candidate exercise set.
The algorithm is shown in \ref{alg:AOA}.
\begin{algorithm}[!h]
    \caption{Hippopotamus Optimization Algorithm}
    \label{alg:AOA}
    \renewcommand{\algorithmicrequire}{\textbf{Input:}}
    \renewcommand{\algorithmicensure}{\textbf{Output:}}
    \begin{algorithmic}[1]
        \REQUIRE $N$, $T$, $d$  
        \ENSURE Exercise recommendation list $L$  
        
        \STATE Initialize population ; evaluate fitness
    \FOR{$t = 1$ to $T$}
        \FOR{$i = 1$ to $N/2$}
            \STATE Compute and update position of $i$-th hippopotamus
        \ENDFOR
        \FOR{$i = N/2+1$ to $N$}
            \STATE Generate predator positions
            \STATE Compute and update position of $i$-th hippopotamus
        \ENDFOR
        \STATE Calculate new decision boundaries
        \FOR{$i = 1$ to $N$}
            \STATE Compute and update position of $i$-th hippopotamus
        \ENDFOR
        
        \STATE Save current best position
    \ENDFOR
    \RETURN Best hippopotamus position
    \end{algorithmic}
\end{algorithm}

\looseness=-1
To achieve a more uniform population distribution in the algorithm, we apply a sine chaotic map aiming at optimizing the initialization phase of HO, generating the initial population $\chi =[\boldsymbol{x}_1,\cdots,\boldsymbol{x}_i,\cdots,\boldsymbol{x}_n]$, where n represents the total number of populations, and each population contains d hippopotamuses. 
The fundamental eqnarray of the Sin chaotic map is as follows:
\begin{eqnarray} 
    \mu _{n+1} = \alpha \sin(\pi / \mu _{n}),
\end{eqnarray}
where the $\mu _{n}$ is the current state value, $\mu_{n+1}$ is the next state value, $\alpha$ a control parameter typically ranging between [0, 1], and  $\pi$ is a constant.

\looseness=-1
In a typical hippopotamus population, there are several female individuals, several juveniles, several male individuals, and one dominant male.
Depending on various circumstances, the members of the population can exhibit different behavioral patterns.
In the hippopotamus optimization algorithm~\cite{AmiMM2024}, these behaviors are classified into three main types: position update in rivers, defense against predators and escaping predators.

\looseness=-1
\textbf{First Behavior}.
Hippopotamuses typically live in groups, dominant males protect the herd’s safety, females gather around them, and adult males may be expelled to find new habitats and form new groups.
The position update of male hippopotamuses in rivers is expressed by the following formula:
\begin{eqnarray} 
    {\omega}_{ij}=x_{ij}+\gamma(B-\eta \cdot x_{ij}),
\end{eqnarray}
where ${\omega}_{ij}$ indicates the location of the male hippopotamus, $B$ denotes the optimal position found so far by the hippopotamus, $\gamma$ is a random number between 0 and 1, and $\eta$ is an integer between 1 and 2.

\looseness=-1
Furthermore, due to their inherent curiosity, juvenile hippopotamuses tend to stray from the group.
However, to protect the juveniles, the female hippopotamuses actively approach them.
Accordingly, their position update is represented by the following formula:
\begin{eqnarray} 
    {\beta }_{ij}=x_{ij}+\gamma(B-\eta \cdot \upsilon_{i}),
\end{eqnarray}
where ${\beta}_{ij}$ represents the positions of the female and juvenile hippopotamuses within the population, $\upsilon_{i}$ refers to the average
of a set of randomly selected hippopotamuses.
We then update the position of the hippopotamus. The formula is as follows:
\begin{eqnarray} 
\boldsymbol{x}_{i} = 
\begin{cases}
\boldsymbol{\omega}_{i}, & \text{if } \rho_{i} < \rho_{\text{best}}, \\
\boldsymbol{x}_{i}, & \text{otherwise},
\end{cases}\quad 
\boldsymbol{x}_{i} = 
\begin{cases}
\boldsymbol{\beta}_{i}, & \text{if } \rho_{i} < \rho_{\text{best}}, \\
\boldsymbol{x}_{i}, & \text{otherwise},
\end{cases}
\end{eqnarray}
where ${\rho}_{i}$ represents the value of fitness function. ${\rho}_{best}$ represents the best value.

\looseness=-1
\textbf{Second Behavior}.
This behavior is inherently defensive and is typically triggered when the animal is attacked by a predator or when another organism invades its territory.
We represent this using the search space of predator as follows:
\begin{eqnarray} 
    \xi_{j}=lb-\gamma(ub-lb),
\end{eqnarray}
where $\xi_{j}$ represents the position of the predator, $lb$ and $ub$ denote the lower and upper bounds of the decision variables.

\looseness=-1
Then, the hippopotamus update its position based on the distance to the predator. The position update of the hippopotamus is represented by the following formula:
\begin{eqnarray} 
\delta_{ij} =
\begin{cases}
l \oplus \xi_{j} + \dfrac{\varphi}{D}, & \text{if } \rho_{\xi} < \rho_{i}, \\
l \oplus \xi_{j} + \dfrac{\varphi}{2D + r_4}, & \text{otherwise},
\end{cases}
\end{eqnarray}
where $\delta _{ij}$ denotes the position of the hippopotamus when facing a predator, and $l$ is a random number following a Lévy distribution, utilized to cause sudden shifts in the predator’s location while attacking the hippopotamus.
The variable $\varphi$ is a uniformly distributed random number between -2 and 2, ${\rho}_{\xi}$ represents the fitness function value of the predator.

\looseness=-1
Next, we update the population. If $\rho_{i}$ is greater than $\rho_{best}$, This implies that the current hippopotamus has been captured and replaced by another one in the group, otherwise, the hunter retreats and the original hippopotamus reenters the herd. The formula is as follows:
\begin{eqnarray} 
\boldsymbol{x}_{i} =
\begin{cases}
\boldsymbol{\delta}_{i}, & \text{if } \rho_{i} < \rho_{best}, \\
\boldsymbol{x}_{i}, & \text{otherwise}.
\end{cases}
\end{eqnarray}

\looseness=-1
\textbf{Third Behavior}.
In this situation, the hippopotamus strives to swim to nearby water bodies, because predators usually avoid entering the water.
To simulate this behavior, we randomly generate a new position near the hippopotamus’s current location.
If the updated position results in a better fitness value, it suggests that the hippopotamus has located a safer surrounding area and modifies its location accordingly.
The position update is expressed by the following formula: 
\begin{equation}
\begin{aligned}
    &ld_{j}=\frac{ld}{t},\ ud_{j}=\frac{ud}{t},\ t=1,2,\dots,T,\\
    &\vartheta_{ij} = x_{ij} + \gamma(ud_{j} - \varphi(ud_{j} - ld_{j})),
\end{aligned}
 \end{equation}
where $\vartheta_{ij}$ represents the the nearest safe position of the hippopotamus. $\varphi$ is a uniformly distributed random number between -1 and 1. Then, update the population position, the formula is as follows:
\begin{eqnarray} 
\boldsymbol{x}_{i} =
\begin{cases}
\boldsymbol{\vartheta}_{i}, & \text{if } \rho_{i} < \rho_{best}, \\
\boldsymbol{x}_{i}, & \text{otherwise}.
\end{cases}
\end{eqnarray}

\looseness=-1
Through these three behavioral patterns, the hippopotamus is able to continually adjust and optimize the population structure, ultimately finding the optimal group distribution.
Therefore, we need to design a fitness function to evaluate the safety of the hippopotamus position, enabling the identification of the optimal position.
Additionally, we design the fitness function as follows:
\begin{eqnarray} 
    \text{Fitness}(\boldsymbol{e}_{i},\boldsymbol{e}_{j})=\sqrt{{(\boldsymbol{e}_{i}-\boldsymbol{e}_j)^2}}.
\end{eqnarray}

\looseness=-1
In the exercise recommendation task, the diversity of exercises is closely related to the fitness value.
A higher fitness value indicates a smaller similarity between two exercises, thus resulting in higher diversity.
In other words, the exercise diversity increases in line with the fitness value.

\section{Experiments}

\subsection{Experiments Settings}
This section mainly introduces the datasets, metrics used and baseline method in the proposed BamaER method.

\noindent\textbf{(i) Dataset.} We experimented with three real-world datasets: ASSISTments2009\footnote{\url{https://sites.google.com/site/assistmentsdata/datasets}}, ASSISTments2012, ASSISTments2017, Algebra2005\footnote{\url{https://pslcdatashop.web.cmu.edu/KDDCup/}} and Bridge2006.
Among them, the first three datasets are derived from the student learning data collected by the ASSITments platform across different academic years, while the latter two datasets are obtained from the KDD Cup 2010 EDM Challenge. All the aforementioned datasets contain core information such as students' exercises and learning interactions.
The detailed information of the datasets is presented in Table \ref{tab:datasets_summary}.

\noindent\textbf{(ii) Metrics.} Considering the aforementioned aspects, a comprehensive evaluation of our algorithm is conducted by employing the following three performance metrics~\cite{LiuRG2025}:
Accuracy: To recommend exercises that match the desired difficulty level for students.
Novelty: To cover knowledge concepts that students have not yet mastered or been exposed to in previous exercises.
Diversity: To ensure the comprehensiveness of knowledge coverage in the recommendation list, thereby aligning with students' actual learning needs.

\noindent\textbf{(iii) Baseline.}
To show the effectiveness of our method, we compared it with the following models: FM~\cite{RenGF2011}, NCF~\cite{HeLZ2017}, SASRec~\cite{KanM2018}, DRER~\cite{HuaLZ2019}, MMER~\cite{LiuHL2023}, LSTMCQP~\cite{HuoWN2020}, KCP-ER~\cite{WuLT2020}, MulOER-SAN~\cite{RenLS2023}, ER-TGA~\cite{LiuRG2025}.

\begin{table}[t]
  \centering
  \caption{Datasets summary. KCs stands for knowledge concepts.}
  \label{tab:datasets_summary}
  \small
  \setlength{\tabcolsep}{3pt} 
  \begin{tabular*}{\linewidth}{@{\extracolsep{\fill}} lrrrr @{}}
    \toprule
    \textbf{Dataset} & \textbf{KCs} & \textbf{Students} & \textbf{Exercises} & \textbf{Records} \\
    \midrule
    ASSISTments2009 & 123 & 4,217 & 26,688 & 346,860 \\
    ASSISTments2012 & 265 & 27,066 & 45,716 & 2,541,201 \\
    ASSISTments2017 & 102 & 1,709 & 2,210 & 942,816 \\
    Algebra2005     & 112 & 574   & 210,710 & 809,694 \\
    Bridge2006      & 493 & 1,146 & 207,856 & 3,679,199 \\
    \bottomrule
  \end{tabular*}
\end{table}

\noindent\textbf{(iv) Implementation Detail.} In the \textit{learning progress prediction}, our model is trained with the Adam optimization method, maintaining a consistent learning rate of 0.0001 throughout.
The training batch size is set to 256. 
In the \textit{knowledge concepts mastery prediction}, the model is trained using the Adam optimization method, with a constant learning rate of 0.0001 and a batch size set to 24, the memory size is set to 20.
In the \textit{exercise filter}, we designate 200 as the candidate set size and limit the recommendation list to 5 items.
For optimization of the recommendation list, we configure the hippopotamus count N as 50, the maximum number of iterations T as 200, and the dimensionality d as 5.
The recommendation list length for all baseline methods is uniformly set to 5.

\begin{table*}[t]
    \centering
    \caption{Comparison of three indicators applying different methods on five datasets.}
    \label{tab:indicators_all}
    \resizebox{\textwidth}{!}{
    \begin{threeparttable}
    \begin{tabular}{c c c c c c c c c c c c c c c c}
        \toprule
        \multirow{2}{*}{\textbf{Methods}} 
        & \multicolumn{3}{c}{\textbf{ASSISTments2009}} 
        & \multicolumn{3}{c}{\textbf{ASSISTments2012}} 
        & \multicolumn{3}{c}{\textbf{ASSISTments2017}} 
        & \multicolumn{3}{c}{\textbf{Algebra2005}} 
        & \multicolumn{3}{c}{\textbf{Bridge2006}} \\
        \cmidrule(lr){2-4} \cmidrule(lr){5-7} \cmidrule(lr){8-10} \cmidrule(lr){11-13} \cmidrule(lr){14-16}
        & Acc & Nov & Div 
        & Acc & Nov & Div 
        & Acc & Nov & Div 
        & Acc & Nov & Div 
        & Acc & Nov & Div \\
        \midrule
        FM & 0.656 & 0.792 & 0.667 & 0.697 & 0.802 & 0.665 & 0.799 & 0.857 & 0.503 & 0.729 & 0.796 & 0.621 & 0.732 & 0.831 & 0.651 \\
        NCF & 0.788 & 0.884 & 0.618 & 0.778 & 0.820 & 0.620 & 0.689 & 0.881 & 0.550 & 0.824 & 0.895 & 0.429 & 0.798 & 0.891 & 0.515 \\
        SASREC & 0.725 & 0.912 & 0.689 & 0.763 & 0.961 & 0.886 & 0.695 & 0.902 & 0.701 & 0.719 & 0.835 & 0.613 & 0.779 & 0.782 & 0.648 \\
        DRER & 0.763 & 0.934 & 0.712 & 0.759 & 0.892 & 0.771 & 0.765 & 0.842 & 0.726 & 0.765 & 0.842 & 0.726 & 0.823 & 0.852 & 0.786 \\
        MMER & 0.728 & 0.951 & 0.693 & 0.753 & 0.911 & 0.721 & 0.790 & 0.912 & 0.703 & 0.750 & 0.811 & 0.641 & 0.776 & 0.832 & 0.698 \\
        LSTMCQP & 0.792 & 0.964 & \underline{0.850} & 0.721 & 0.933 & 0.823 & 0.933 & 0.981 & 0.897 & \textbf{0.956} & 0.851 & 0.722 & 0.863 & 0.852 & 0.748 \\
        KCP-ER & 0.899 & \underline{0.995} & 0.827 & \textbf{0.966} & 0.982 & 0.899 & 0.925 & 0.986 & 0.964 & 0.866 & 0.839 & 0.851 & \underline{0.960} & 0.927 & 0.807 \\
        MulOER-SAN & 0.908 & 0.988 & 0.841 & 0.941 & \underline{0.993} & 0.921 & \underline{0.978} & 0.983 & 0.971 & 0.894 & 0.903 & 0.898 & 0.947 & 0.923 & 0.818 \\
        ER-TGA & \underline{0.921} & 0.989 & 0.849 & 0.923 & 0.991 & \underline{0.964} & 0.975 & \underline{0.991} & \underline{0.976} & 0.901 & \underline{0.941} & \underline{0.918} & 0.958 & \underline{0.945} & \underline{0.889}\\
        \midrule 
        BamaER & \textbf{0.960} & \textbf{0.997} & \textbf{0.901} & \underline{0.955} & \textbf{0.999} & \textbf{0.985} & \textbf{0.991} & \textbf{0.998} & \textbf{0.980} & \underline{0.931} & \textbf{0.999} & \textbf{0.948} & \textbf{0.983} & \textbf{0.996} & \textbf{0.938} \\
        \bottomrule
    \end{tabular}
    \end{threeparttable}
    }
\end{table*}

\begin{table*}[t]
    \centering
    \caption{Comparison of ablation experiments on five datasets.}
    \label{tab:indicators_bamaer}
    \resizebox{\textwidth}{!}{
    \begin{tabular}{l ccc ccc ccc ccc ccc}  
        \toprule
        \multirow{2}{*}{\textbf{Methods}} 
        & \multicolumn{3}{c}{\textbf{ASSISTments2009}} 
        & \multicolumn{3}{c}{\textbf{ASSISTments2012}} 
        & \multicolumn{3}{c}{\textbf{ASSISTments2017}} 
        & \multicolumn{3}{c}{\textbf{Algebra2005}} 
        & \multicolumn{3}{c}{\textbf{Bridge2006}} \\
        \cmidrule(lr){2-4} \cmidrule(lr){5-7} \cmidrule(lr){8-10} \cmidrule(lr){11-13} \cmidrule(lr){14-16}
        & Acc & Nov & Div 
        & Acc & Nov & Div 
        & Acc & Nov & Div 
        & Acc & Nov & Div 
        & Acc & Nov & Div \\
        \midrule
        BamaER     & \textbf{0.960} & \textbf{0.997} & \textbf{0.901} & \textbf{0.955} & \textbf{0.999} & \textbf{0.985} & \textbf{0.991} & \textbf{0.998} & \textbf{0.980} & \textbf{0.931} & \textbf{0.999} & \textbf{0.948} & \textbf{0.983} & \textbf{0.996} & \textbf{0.938} \\
        BamaER w/o MAKT   & 0.918 & 0.989 & 0.873 & 0.873 & 0.991 & 0.953 & 0.941 & 0.997 & 0.959 & 0.895 & 0.981 & 0.908 & 0.969 & 0.991 & 0.912 \\
        BamaER w/o LP  & 0.917 & 0.976 & 0.871  & 0.925 & 0.987 & 0.943  & 0.934 & 0.995 & 0.970 & 0.915 & 0.978 & 0.915 & 0.969 & 0.994 & 0.922 \\
        BamaER w/o ER-HO & 0.932 & 0.987 & 0.755 & 0.930 & 0.989 & 0.834 & 0.936 & 0.997 & 0.881 & 0.916 & 0.989 & 0.726 & 0.972 & 0.993 & 0.732 \\
        \bottomrule
    \end{tabular}
    }
\end{table*}

\begin{figure}[h]
  \centering
  \includegraphics[width=0.48\textwidth]{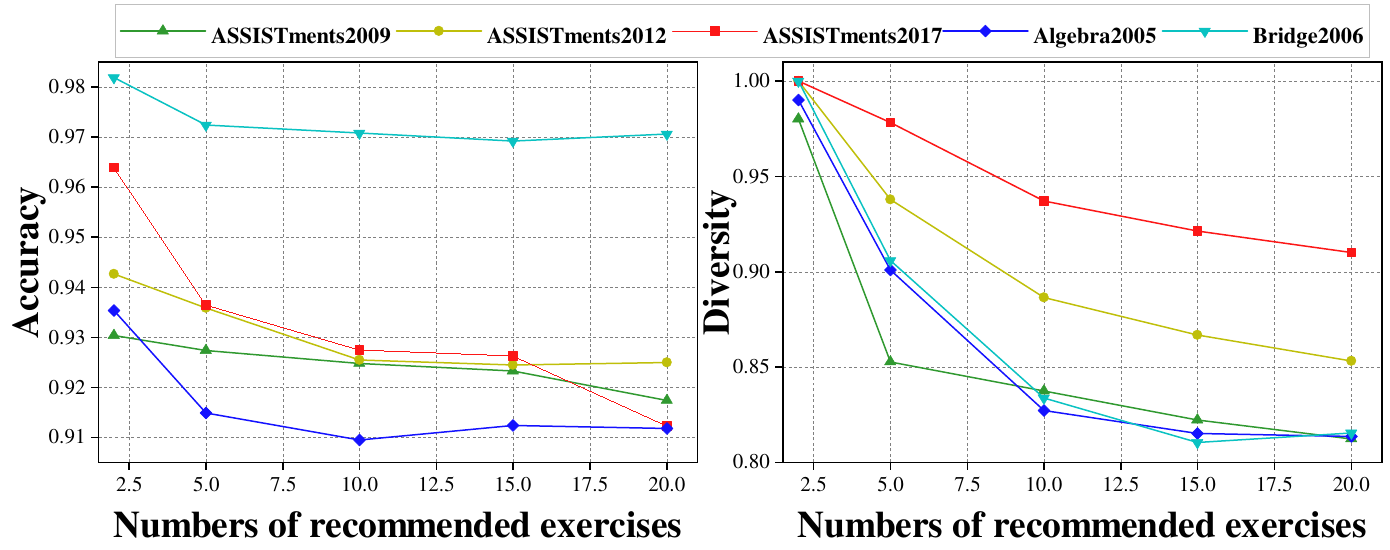} 
  \caption{Comparison of accuracy and diversity metrics under different numbers of exercise recommendations on five datasets.}
  \label{fig:diffrent_number_of_ex}
\end{figure}

\begin{figure}[h]
  \centering
  \includegraphics[width=0.48\textwidth]{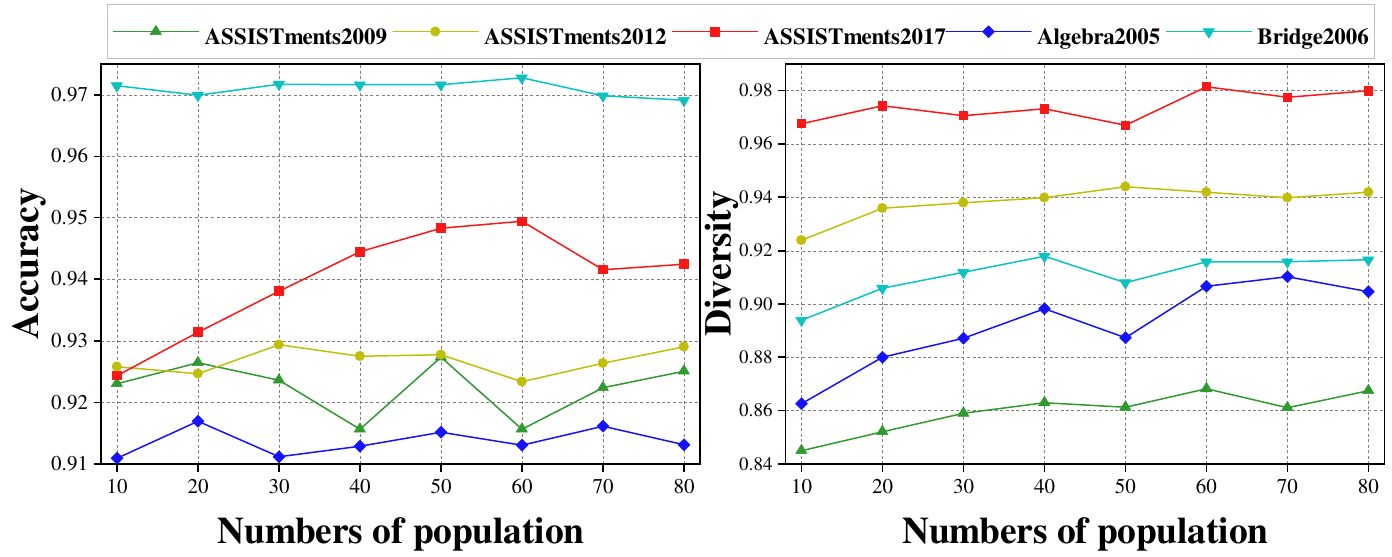} 
  \caption{Comparison of accuracy and diversity metrics under different numbers of population on five datasets.}
  \label{fig:Numbers_of_population}
\end{figure}

\begin{figure}[h]
	\centering
	\includegraphics[width=0.23\textwidth]{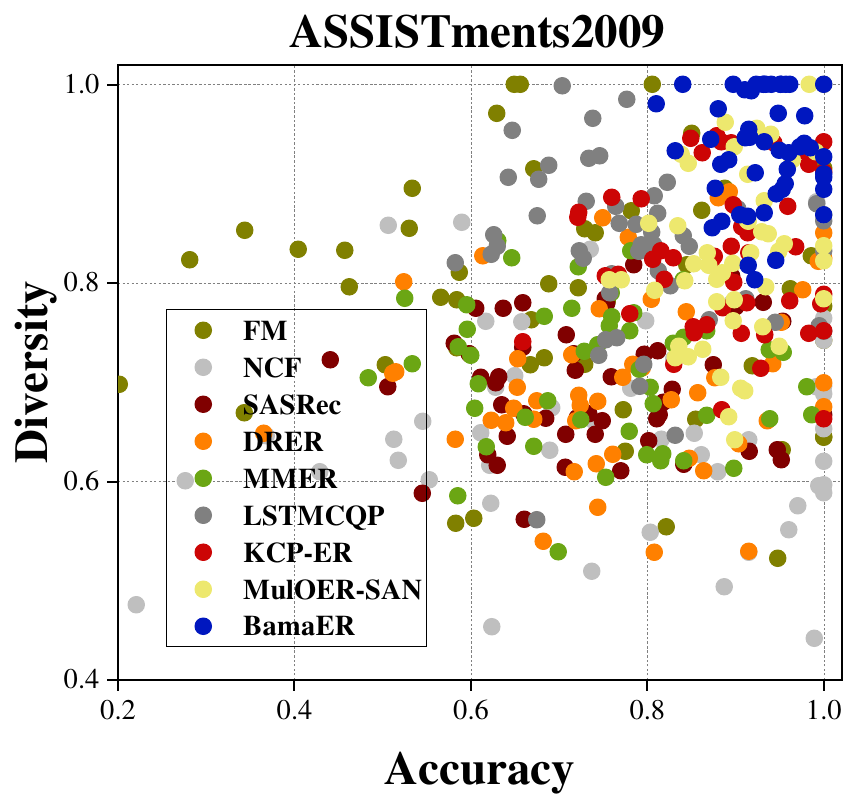}
	\includegraphics[width=0.23\textwidth]{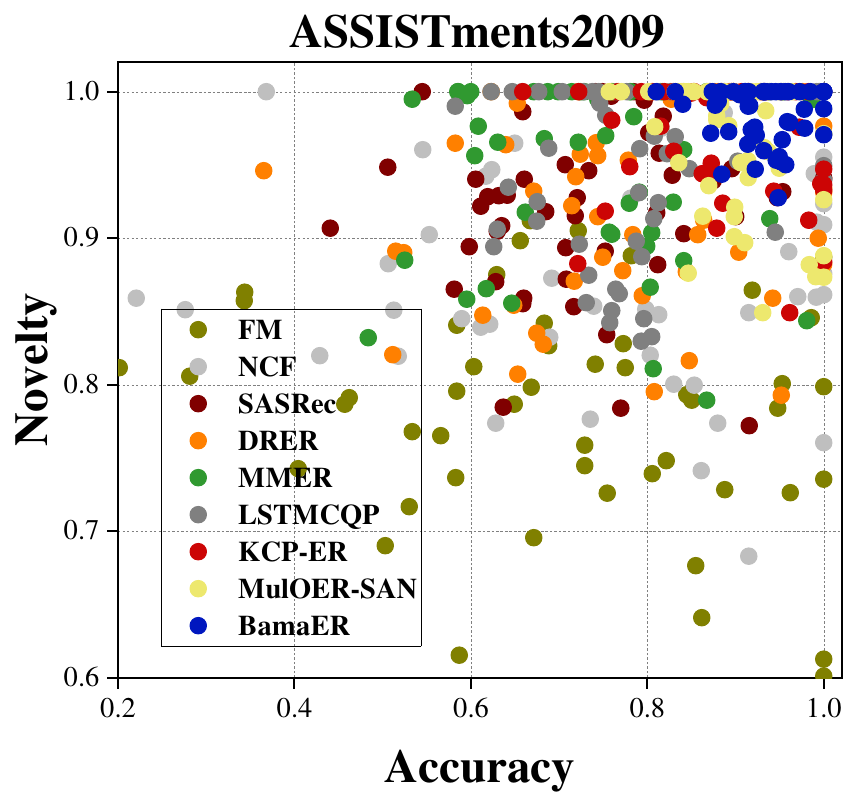}
    \includegraphics[width=0.23\textwidth]{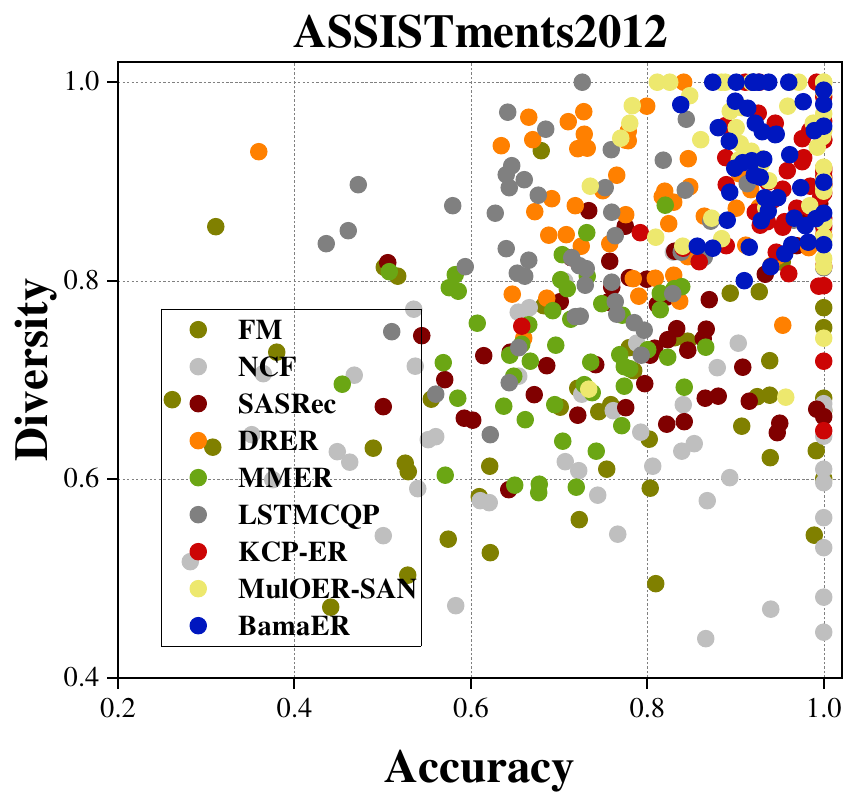} 
    \includegraphics[width=0.23\textwidth]{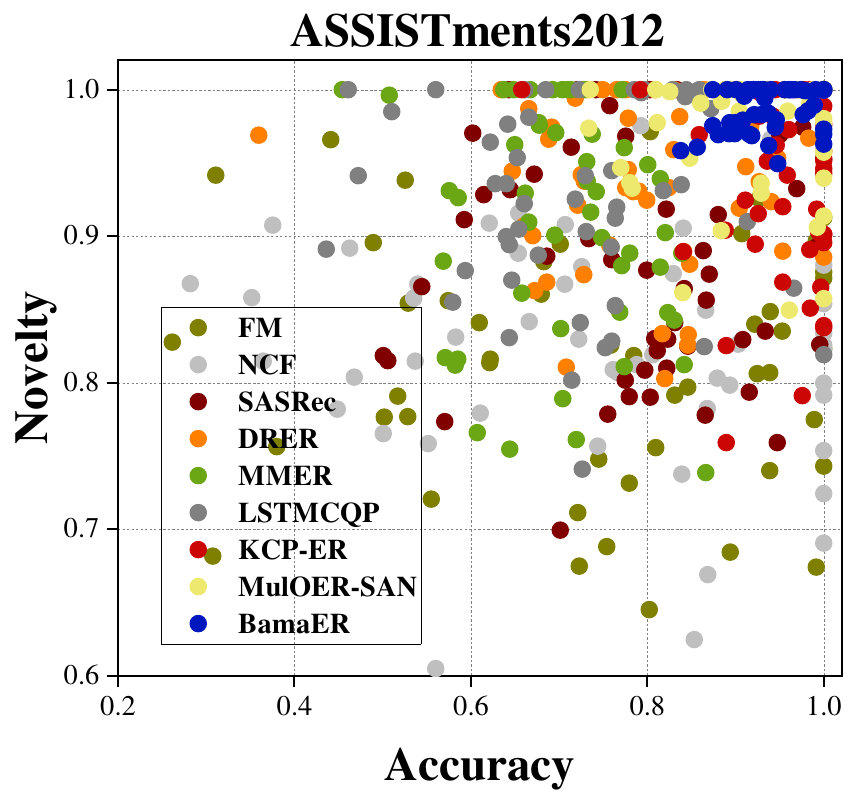}
	\includegraphics[width=0.23\textwidth]{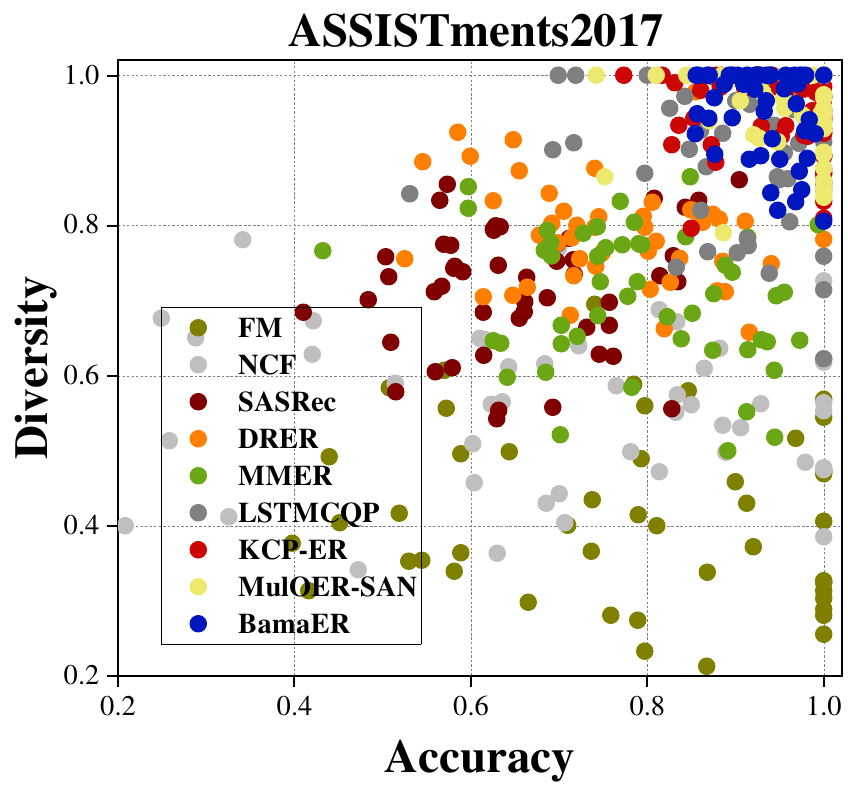}
    \includegraphics[width=0.23\textwidth]{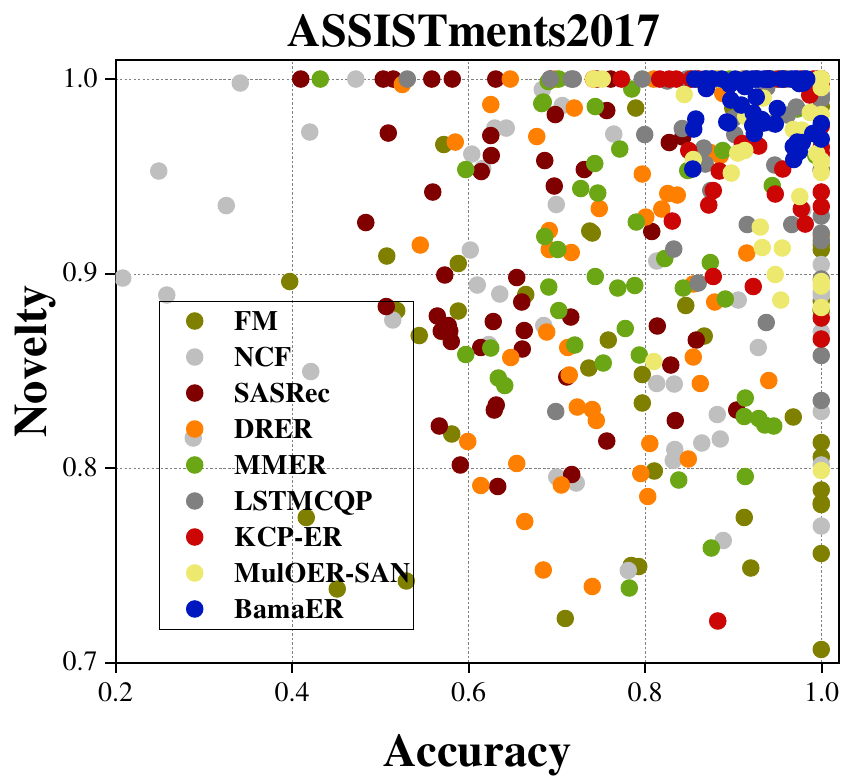}
    \includegraphics[width=0.23\textwidth]{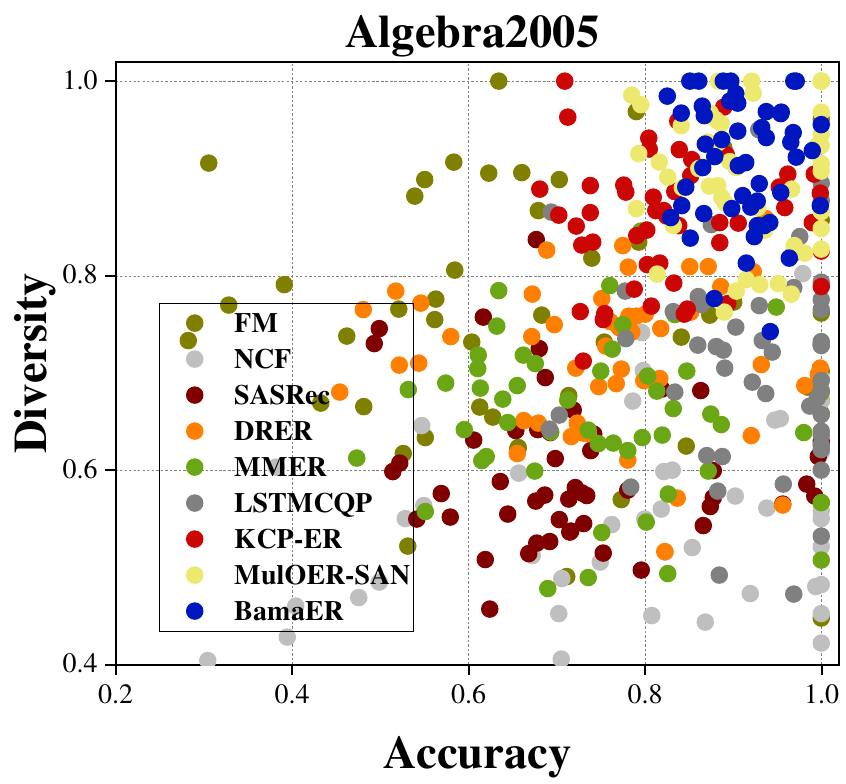}
    \includegraphics[width=0.23\textwidth]{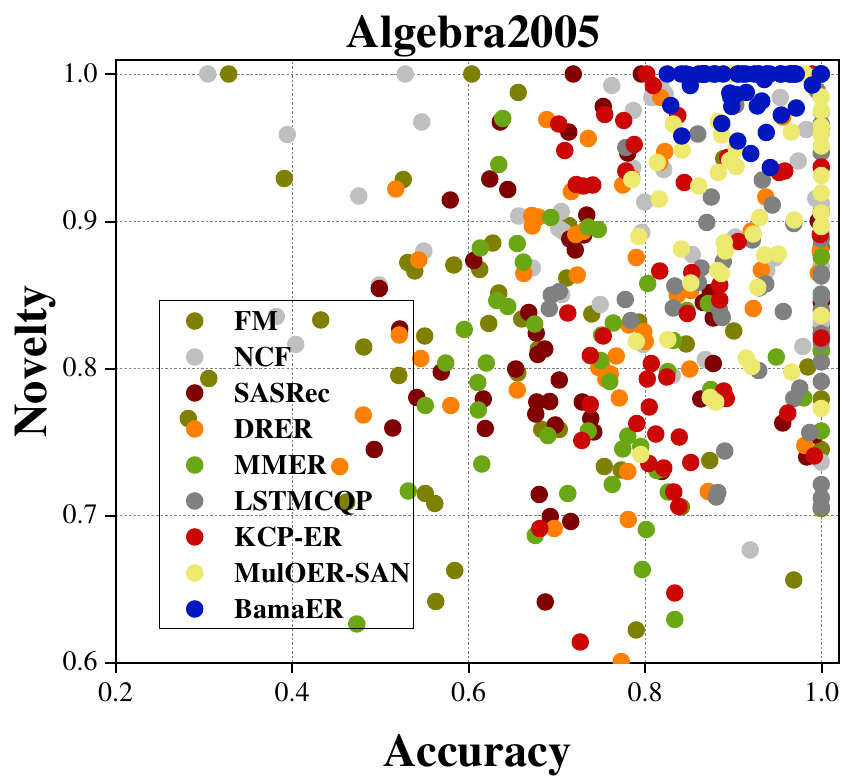}
    \includegraphics[width=0.23\textwidth]{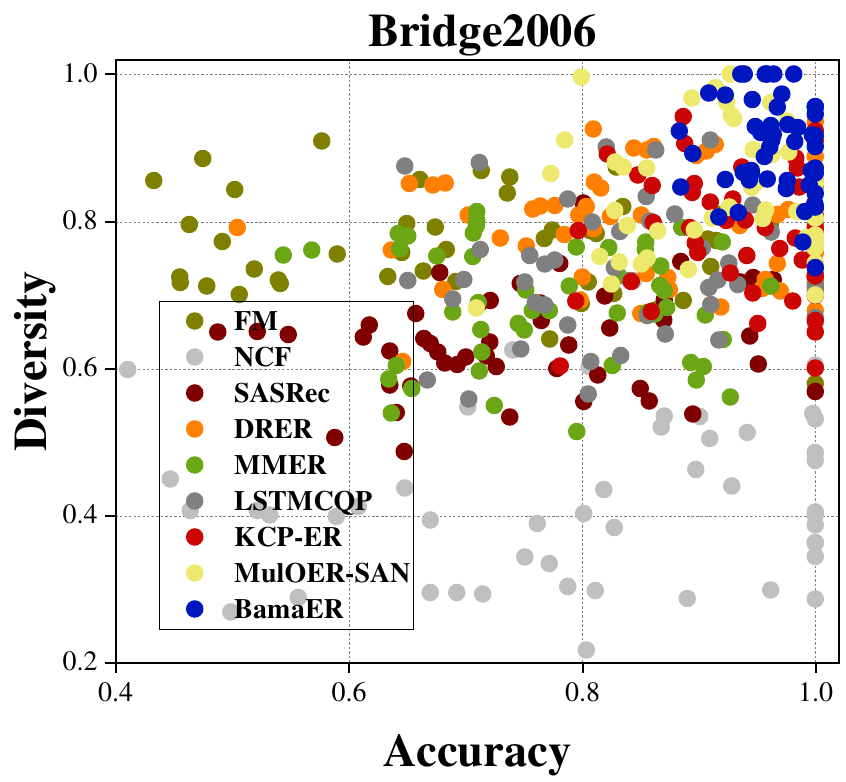}
    \includegraphics[width=0.23\textwidth]{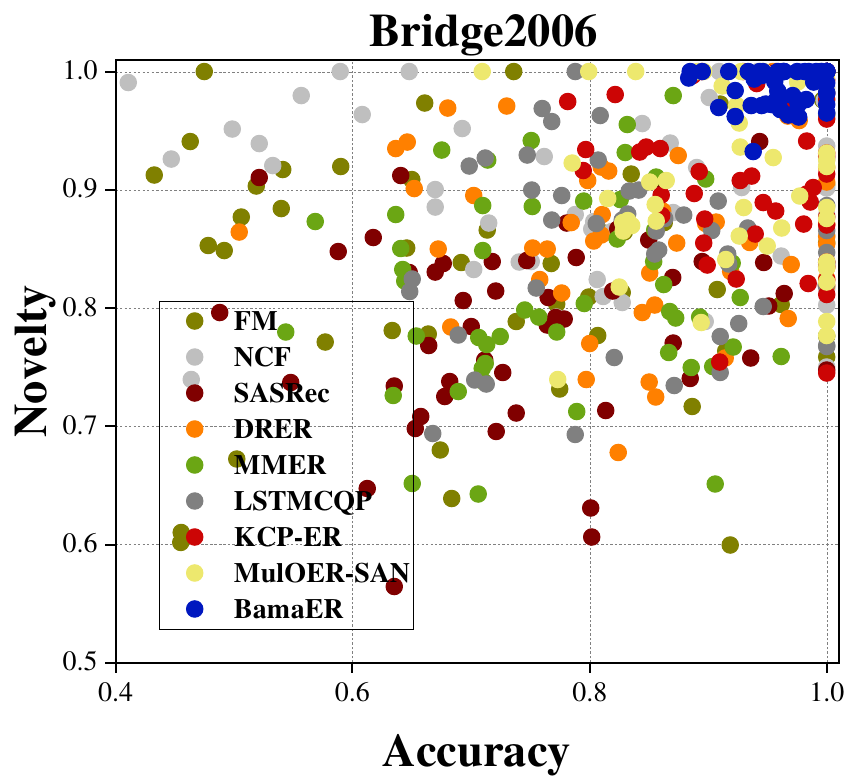}
	\caption{Detailed comparison of the performance of the exercise recommendation list across three key metrics on ASSISTments datasets.}
	\label{fig:Banlance}
\end{figure}



\subsection{Performance Comparison}
To verify the effectiveness of the BamaER model, we compared BamaER with the baseline methods on five datasets.
In addition, the values of each metric obtained from the experiments represent the average results from multiple runs, with higher values indicating better performance, thereby ensuring the reliability and reproducibility of the findings.
The experimental analyses are list as follws:

(1) As shown in Table \ref{tab:indicators_all}, the bolded values represent the best results, while the underlined values represent the second-best results.
The results show that BamaER outperforms the baselines on multiple datasets, demonstrating the effectiveness of our method.
By achieving higher scores on key metrics, BamaER exhibits outstanding overall performance in complex exercise recommendation tasks.

\looseness = -1
(2) In the experiments, collaborative filtering-based methods performed poorly overall, primarily because they rely solely on exercise similarity for recommendations.
This approach fails to capture knowledge states of students and individual differences, resulting in limited recommendation accuracy.

(3) Among the machine learning-based methods, SASREC and MMER also exhibited relatively weaker performance.
SASREC, being a purely sequential recommendation model, lacks the capacity to model learning goals and knowledge mastery of students. MMER, by contrast, generates recommendations solely based on competition among knowledge concepts, without considering multiple learning objectives.
These limitations contribute to their underperformance across several evaluation metrics.

(4) In contrast, the four baseline models (LSTMCQP, KCP-ER, MulOER-SAN, and ER-TGA) perform relatively well. Furthermore, KCP-ER and LSTMCQP outperform BamaER on certain metrics across specific datasets, indicating that they possess advantages in specific tasks or evaluation dimensions.
This is because these methods also incorporate multiple learning metrics and explore a variety of semantic relationships, which may enhance their performance on specific evaluation criteria.

Despite these differences, our method still demonstrates significant advantages on other key metrics, proving its outstanding capability in balancing various recommendation objectives.
This is mainly due to our integration of complex learner interaction information, which improves the accuracy of learning progress prediction. Moreover, the memory-enhanced knowledge tracing module effectively prevents the fragmentation of early learning experiences.
By improving the modeling of both learning progress and mastery, our method enables a more comprehensive understanding of knowledge states, leading to more accurate and personalized exercise recommendations.

\subsection{Ablation Experiments}

In this section, we will conduct detailed ablation experiments on the BameER model to explore the specific impact of each module on the model's performance. We will perform experiments on three model variants, each with one key component of the model removed. The following is a detailed description of these variants:

\begin{itemize}[left=0cm]
    \item BamaER w/o MAKT removes the knowledge concept mastery prediction module, without accounting for the impact of knowledge concept mastery.
    \item BamaER w/o LP removes the learning progress prediction module, disregarding the impact of students' learning progress.
    \item BamaER w/o ER-HO removes the algorithm optimization module, instead randomly selecting exercises from the candidate set.
\end{itemize}

\looseness=-1
As shown in the Table \ref{tab:indicators_bamaer}, removing different modules affects all performance metrics of the model to varying degrees:

(1) Specifically, BamaER w/o MAKT and BamaER w/o LP primarily compromise the model’s overall accuracy, as these two variants correspond to students' knowledge concept mastery and learning progress, respectively. Inability to accurately predict such information results in recommended exercises lacking relevance and misaligning with students' current learning status, and this aligns with our original objective of developing a model capable of accurately predicting students' learning status.

\looseness=-1
(2) By comparison, after removing the ER-HO module, the diversity metric of BamaER w/o ER-HO decreases notably. This indicates the presence of numerous redundant items in the candidate set, and the ER-HO module can effectively identify and eliminate these redundancies, thereby enhancing the model’s overall diversity.

\looseness=-1
(3) Furthermore, Table \ref{tab:indicators_bamaer} shows that the three model variants have only a slight negative effect on novelty. This is because novelty depends largely on the candidate set screening method and the number of knowledge concepts in the exercise set—and the candidate set’s novelty generally resembles that of the final recommendations.

\subsection{Parameter Sensitivity Analysis}
In this subsection, we analyze the impact of parameter settings on the accuracy and diversity metrics to ensure the robustness and reproducibility of BamaER. The parameters examined include the number of recommended exercises and the population size.

\looseness=-1
\textbf{Number of Recommended Exercises.} For a student, recommending an appropriate number of exercises is crucial. We tested five different settings: 2, 5, 10, 15, and 20.
As shown in the Figure \ref{fig:diffrent_number_of_ex}, both accuracy and diversity reached their highest levels when only two exercises were recommended.
However, as the number of recommended exercises increased, both accuracy and diversity gradually decreased.
Considering that recommending only two exercises is insufficient to meet practice needs of students, we ultimately set the number of recommended exercises to 5.

\looseness=-1
\textbf{Population Size.} Population size is a key parameter in the HO. We tested eight different settings: 10, 20, 30, 40, 50, 60, 70, and 80.
As illustrated in the Figure \ref{fig:Numbers_of_population}, the population size has a minimal impact on the accuracy and diversity of the recommended exercise list.
After comprehensively considering both accuracy and diversity, we set the population size parameter to 50.

\subsection{Analysis of Balancing among Accuracy, Novelty and Diversity}

To examine whether our approach achieves an effective balance between accuracy and serendipity (i.e., novelty and diversity), we employ scatter plots to analyze the relationship between accuracy and each serendipity metric.
As illustrated in Figure~\ref{fig:Banlance}, the x-axis denotes recommendation accuracy, while the y-axis corresponds to diversity or novelty, respectively.
For robustness, we randomly sample 50 students from each dataset and generate user-level scatter plots, enabling a fine-grained assessment of individual recommendation performance.

In these plots, the upper-right region represents the ideal operating regime, where high accuracy and high serendipity are simultaneously achieved.
Compared to competing methods, the scatter points produced by BamaER are consistently closer to this ideal region and exhibit a noticeably denser distribution.
This pattern indicates not only superior average performance but also reduced variance across individual users.
In contrast to methods that achieve high accuracy or high serendipity sporadically, BamaER demonstrates a more stable and consistent trade-off between the two objectives.
Overall, these observations confirm that BamaER effectively balances accuracy and serendipity, maintaining a favorable operating regime relative to baseline approaches and further validating the effectiveness of the proposed framework.

\section{Conclusion}
This paper proposes a \underline{B}ehavior-\underline{a}ware \underline{m}emory-\underline{a}ugmented \underline{E}xercise \underline{R}ecommendation (BamaER) framework composed of three core modules: (i) the learning progress prediction module, which incorporates rich behavioral interaction information from students’ historical response sequences, going beyond simple exercise order; (ii) the knowledge concept mastery prediction module, which maintains a memory matrix to preserve early-stage knowledge states and jointly model them with current learning states, enabling the representation of both long-term retention and short-term learning dynamics for more accurate and stable mastery estimation; and (iii) the exercise filter module, which leverages predictions from the preceding two modules to construct candidate exercise sets and applies the Hippopotamus Optimization Algorithm to enhance recommendation diversity.
Extensive experiments on five real-world datasets demonstrate the effectiveness of each component and show that BamaER consistently outperforms eight representative baseline methods.



\bibliographystyle{ACM-Reference-Format}
\bibliography{refs}

@STRING{jun = "June"}

@article{WuLT2020,
  title={Exercise recommendation based on knowledge concept prediction},
  author={Wu, Zhengyang and Li, Ming and Tang, Yong and Liang, Qingyu},
  journal={Knowledge-Based Syst.},
  volume={210},
  pages={106481},
  year={2020}
}

@article{RenLS2023,
  author={Ren, Yimeng and Liang, Kun and Shang, Yuhu and Zhang, Yiying},
  title={MulOER-SAN: 2-layer multi-objective framework for exercise recommendation with self-attention networks},
  journal={Knowledge-Based Syst.},
  volume={260},
  pages={110117},
  year={2023}
}

@inproceedings{LiuHL2023,
  title={Meta Multi-agent Exercise Recommendation: A Game Application Perspective},
  author={Liu, Fei and Hu, Xuegang and Liu, Shuochen and Bu, Chenyang and Wu, Le},
  booktitle={Proc. 29th ACM SIGKDD Int. Conf. Knowl. Discov. Data Min. (SIGKDD)},
  address={New York, NY, USA},
  pages={1441-1452},
  year={2023}
}

@inproceedings{PieBH2015,
  author={Piech, Chris and Bassen, Jonathan and Huang, Jonathan and Ganguli, Surya and Sahami, Mehran and Guibas, Leonidas J and Sohl-Dickstein, Jascha},
  title={Deep knowledge tracing},
  booktitle={Proc. Adv. neural inf. proces. syst. (NeurIPS)},
  address={Montreal, QC, Canada},
  year={2015}
}

@inproceedings{GuaXC2023,
  author={Guan, Quanlong and Xiao, Fang and Cheng, Xinghe and Fang, Liangda and Chen, Ziliang and Chen, Guanliang and Luo, Weiqi},
  title={KG4Ex: An Explainable Knowledge Graph-Based Approach for Exercise Recommendation},
  booktitle={Proc. 32nd Int Conf Inf Knowledge Manage (CIKM)},
  address={Birmingham, UK},
  pages={597-607},
  year={2023}
}

@inproceedings{HuaLZ2019,
  author={Huang, Zhenya and Liu, Qi and Zhai, Chengxiang and Yin, Yu and Chen, Enhong and Gao, Weibo and Hu, Guoping},
  title={Exploring Multi-Objective Exercise Recommendations in Online Education Systems},
  booktitle={Proc. 28th Int Conf Inf Knowledge Manage (CIKM)},
  address={Beijing, China},
  pages={1261-1270},
  year={2019}
}

@article{LiuRG2025,
  title={Comprehensive exercise recommendation with practicality, generalizability, and versatility in AI-driven education},
  author={Liu, Guowei and Ren, Meirui and Guo, Longjiang and Li, Jin and Ma, Miao},
  journal={Inf. Process. Manage.},
  volume={62},
  number={3},
  pages={104051},
  year={2025}
}

@inproceedings{GhoHL2020,
  author={Ghosh, Aritra and Heffernan, Neil and Lan, Andrew S},
  title={Context-Aware Attentive Knowledge Tracing},
  booktitle={Proc. 26th ACM SIGKDD Int. Conf. Knowl. Discov. Data Min (SIGKDD-2020)},
  address={Online, USA},
  pages={2330-2339},
  year={2020}
}

@inproceedings{LiuWB2022,
  author={Liu, Naiming and Wang, Zichao and Baraniuk, Richard and Lan, Andrew},
  title={Open-Ended Knowledge Tracing for Computer Science Education},
  booktitle={Proc. Conf. Empir. Methods Nat. Lang. Process. (EMNLP)},
  address={Abu Dhabi, United arab emirates},
  year={2022}
}

@article{AbdWN2023,
  author={Abdelrahman, Ghodai and Wang, Qing and Nunes, Bernardo},
  title={Knowledge Tracing: A Survey},
  journal={ACM Comput. Surv.},
  volume={55},
  number={11},
  pages={1-37},
  year={2023}
}

@article{HuoWN2020,
  author={Huo, Yujia and Wong, Derek F and Ni, Lionel M and Chao, Lidia S and Zhang, Jing},
  title={Knowledge modeling via contextualized representations for LSTM-based personalized exercise recommendation},
  journal={Inf. Sci.},
  volume={523},
  pages={266-278},
  year={2020},
  publisher={Elsevier}
}

@inproceedings{WanDR2006,
  author={Wang, Jun and De Vries, Arjen P and Reinders, Marcel JT},
  title={Unifying user-based and item-based collaborative filtering approaches by similarity fusion},
  booktitle={Proc. 29th Annu. Int. ACM SIGIR Conf. Res. Dev. Inf. Retr.(SIGIR)},
  address={Seattle, WA, USA},
  pages={501-508},
  year={2006}
}

@inproceedings{SarKK2001,
  author={Sarwar, Badrul and Karypis, George and Konstan, Joseph and Riedl, John},
  title={Item-based collaborative filtering recommendation algorithms},
  booktitle={Proc. 10th Int. Conf. World Wide Web (WWW)},
  address={Hong Kong},
  pages={285-295},
  year={2001}
}

@inproceedings{HeLZ2017,
  author={He, Xiangnan and Liao, Lizi and Zhang, Hanwang and Nie, Liqiang and Hu, Xia and Chua, Tat-Seng},
  title={Neural Collaborative Filtering},
  booktitle={Proc. 26th Int. World Wide Web Conf. (WWW)},
  address={Perth, WA, Australia},
  pages={173-182},
  year={2017}
}

@inproceedings{KanM2018,
  author={Kang, Wang-Cheng and McAuley, Julian},
  title={Self-Attentive Sequential Recommendation},
  booktitle={Proc. IEEE Int. Conf. Data Min. (ICDM)},
  address={Pisa, Italy},
  pages={197-206},
  year={2018}
}

@inproceedings{RenGF2011,
  author={Rendle, Steffen and Gantner, Zeno and Freudenthaler, Christoph and Schmidt-Thieme, Lars},
  title={Fast context-aware recommendations with factorization machines},
  booktitle={Proc. 34th Proc. Int. ACM SIGIR Conf. Res. Dev. Inf. Retr. (SIGIR)},
  address={Beijing, China},
  pages={635-644},
  year={2011}
}

@article{AmiMM2024,
  author={Amiri, Mohammad Hussein and Mehrabi Hashjin, Nastaran and Montazeri, Mohsen and Mirjalili, Seyedali and Khodadadi, Nima},
  title={Hippopotamus optimization algorithm: a novel nature-inspired optimization algorithm},
  journal={Sci. Rep.},
  volume={14},
  number={1},
  pages={5032},
  year={2024}
}

@article{HenG2016,
  author={Hendrycks, Dan and Gimpel, Kevin},
  title={Gaussian Error Linear Units (GELUs)},
  journal={arXiv preprint arXiv:1606.08415},
  year={2016}
}

@article{BaKH2016,
  author={Ba, Jimmy Lei and Kiros, Jamie Ryan and Hinton, Geoffrey E},
  title={Layer Normalization},
  journal={arXiv preprint arXiv:1607.06450},
  year={2016}
}

@article{ElfUD2018,
  author={Elfwing, Stefan and Uchibe, Eiji and Doya, Kenji},
  title={Sigmoid-weighted linear units for neural network function approximation in reinforcement learning},
  journal={Neural Networks},
  volume={107},
  pages={3-11},
  year={2018}
}

@article{NwaIG2018,
  author={Nwankpa, Chigozie and Ijomah, Winifred and Gachagan, Anthony and Marshall, Stephen},
  title={Activation Functions: Comparison of trends in Practice and Research for Deep Learning},
  journal={arXiv preprint arXiv:1811.03378},
  year={2018}
}

@article{GaoZL2024,
  author={Gao, Jingtong and Zhao, Xiangyu and Li, Muyang and Zhao, Minghao and Wu, Runze and Guo, Ruocheng and Liu, Yiding and Yin, Dawei},
  title={SMLP4Rec: An Efficient All-MLP Architecture for Sequential Recommendations},
  journal={ACM Trans. Inf. Syst.},
  volume={42},
  number={3},
  pages={1-23},
  year={2024}
}

@inproceedings{CheKH2016,
  author={Cheng, Heng-Tze and Koc, Levent and Harmsen, Jeremiah and Shaked, Tal and Chandra, Tushar and Aradhye, Hrishi and Anderson, Glen and Corrado, Greg and Chai, Wei and Ispir, Mustafa and others},
  title={Wide \& Deep Learning for Recommender Systems},
  booktitle={Proc. 1st ACM Int. Conf. Proc. Ser. (DLRS)},
  pages={7-10},
  year={2016}
}

@article{HeLW2023,
  author={He, Liangliang and Li, Xiao and Wang, Pancheng and Tang, Jintao and Wang, Ting},
  title={MAN: Memory-augmented Attentive Networks for Deep Learning-based Knowledge Tracing},
  journal={ACM Trans. Inf. Syst.},
  volume={42},
  number={1},
  pages={1-22},
  year={2023}
}

@inproceedings{PazB2007,
  title={Content-Based Recommendation Systems},
  author={Pazzani, Michael J and Billsus, Daniel},
  booktitle={The Adaptive Web: Methods and Strategies of Web Personalization},
  pages={325-341},
  address={Boston, MA, USA},
  year={2007}
}

@article{WalRL2004,
  author={Walker, Andrew and Recker, Mimi M and Lawless, Kimberly and Wiley, David},
  title={Collaborative Information Filtering: A Review and an Educational Application},
  journal={Int. J. Artif. Intell. Educ.},
  volume={14},
  number={1},
  pages={3-28},
  year={2004}
}

@article{ChaLC2016,
  author={Chang, Pei-Chann and Lin, Cheng-Hui and Chen, Meng-Hui},
  title={A Hybrid Course Recommendation System by Integrating Collaborative Filtering and Artificial Immune Systems},
  journal={Algorithms},
  volume={9},
  number={3},
  pages={47},
  year={2016}
}

@article{KlaIN2015,
  author={Kla{\v{s}}nja-Mili{\'c}evi{\'c}, Aleksandra and Ivanovi{\'c}, Mirjana and Nanopoulos, Alexandros},
  title={Recommender systems in e-learning environments: a survey of the state-of-the-art and possible extensions},
  journal={Artif. Intell. Rev.},
  volume={44},
  pages={571-604},
  year={2015}
}

@article{GeoL2019,
  author={George, Gina and Lal, Anisha M},
  title={Review of ontology-based recommender systems in e-learning},
  journal={Comput Educ.},
  volume={142},
  pages={103642},
  year={2019}
}

@article{GraG2012,
  title={Long Short-Term Memory},
  author={Graves, Alex and Graves, Alex},
  journal={Supervised Sequence Labelling with Recurrent Neural Networks},
  pages={37-45},
  year={2012}
}

@article{SonLC202,
  author={Song, Xiangyu and Li, Jianxin and Cai, Taotao and Yang, Shuiqiao and Yang, Tingting and Liu, Chengfei},
  title={A survey on deep learning based knowledge tracing},
  journal={Knowledge-Based Syst.},
  volume={258},
  pages={110036},
  year={2022}
}

@inproceedings{VasSP2017,
 author = {Vaswani, Ashish and Shazeer, Noam and Parmar, Niki and Uszkoreit, Jakob and Jones, Llion and Gomez, Aidan N and Kaiser, \L ukasz and Polosukhin, Illia},
 booktitle = {Proc. Adv. neural inf. proces. syst. (NeurIPS)},
 address={Long Beach, CA, USA},
 title = {Attention is All you Need},
 year = {2017}
}

@inproceedings{ZhaSK2017,
  author={Zhang, Jiani and Shi, Xingjian and King, Irwin and Yeung, Dit-Yan},
  title={Dynamic Key-Value Memory Networks for Knowledge Tracing},
  booktitle={Proc. 26th Int. World Wide Web Conf. (WWW)},
  address={Perth, WA, Australia},
  pages={765-774},
  year={2017}
}

@article{PanK2019,
  title={A Self-Attentive model for Knowledge Tracing},
  author={Pandey, Shalini and Karypis, George},
  journal={arXiv preprint arXiv:1907.06837},
  year={2019}
}

@inproceedings{SanBB2016,
  title={Meta-Learning with Memory-Augmented Neural Networks},
  author={Santoro, Adam and Bartunov, Sergey and Botvinick, Matthew and Wierstra, Daan and Lillicrap, Timothy},
  booktitle={Proc. 33rd Int. ACM Conf. Knowl. Mac. Lea. (PMLR)},
  address={New York NY USA},
  pages={1842--1850},
  year={2016}
}

@inproceedings{SukWF2015,
  title={End-To-End Memory Networks},
  author={Sukhbaatar, Sainbayar and Weston, Jason and Fergus, Rob and others},
  journal={Proc. Adv. neural inf. proces. syst. (NeurIPS)},
  address={Montreal, QC, Canada},
  volume={28},
  year={2015}
}

@inproceedings{WanKL2024,
  title={Large Language Models-guided Dynamic Adaptation for Temporal Knowledge Graph Reasoning},
  author={Wang, Jiapu and Kai, Sun and Luo, Linhao and Wei, Wei and Hu, Yongli and Liew, Alan Wee-Chung and Pan, Shirui and Yin, Baocai},
  journal={Proc. Adv. neural inf. proces. syst. (NeurIPS)},
  address={Montreal, QC, Canada},
  pages={8384-8410},
  year={2024}
}

@article{WanWG2024,
  title={MADE: Multicurvature Adaptive Embedding for Temporal Knowledge Graph Completion},
  author={Wang, Jiapu and Wang, Boyue and Gao, Junbin and Pan, Shirui and Liu, Tengfei and Yin, Baocai and Gao, Wen},
  journal={IEEE Trans. Cybern.},
  volume={54},
  number={10},
  pages={5818-5831},
  year={2024}
}

@inproceedings{WanCW2024,
  title={IME: Integrating multi-curvature shared and specific embedding for temporal knowledge graph completion},
  author={Wang, Jiapu and Cui, Zheng and Wang, Boyue and Pan, Shirui and Gao, Junbin and Yin, Baocai and Gao, Wen},
  booktitle={Proc. Companion Proc. ACM Web Conf. (WWW)},
  address={Singapore, Singapore},
  pages={1954--1962},
  year={2024}
}

@inproceedings{CheZF2025,
  title={NR4DER: Neural Re-ranking for Diversified Exercise Recommendation},
  author={Cheng, Xinghe and Zhou, Xufang and Fang, Liangda and He, Chaobo and Zhou, Yuyu and Luo, Weiqi and Gong, Zhiguo and Guan, Quanlong},
  booktitle={arXiv:2506.06341},
  year={2025}
}

@article{CheGX2025,
  title={Explainable exercise recommendation with knowledge graph},
  author={Guan, Quanlong and Cheng, Xinghe and Xiao, Fang and Li, Zhuzhou and He, Chaobo and Fang, Liangda and Chen, Guanliang and Gong, Zhiguo and Luo, Weiqi},
  journal={Neural Networks},
  volume={183},
  pages={106954},
  year={2025}
}

@article{SheYL2024,
  author={Shen, Xiaoxuan and Yu, Jianwei and Liang, Ruxia and Li, Qing and Liu, Shengyingjie and Du, Shangheng and Sun, Jianwen and Liu, Sannyuya},
  journal={IEEE Trans. Neural Networks Learn. Syst.}, 
  title={Autobalanced Multitask Node Embedding Framework for Intelligent Education}, 
  year={2024},
  volume={35},
  number={6},
  pages={8653-8667}}

@article{LiDQ2025,
  author={Li, Jiawei and Deng, Yuanfei and Qin, Yixiu and Mao, Shun and Jiang, Yuncheng},
  journal={IEEE Trans. Neural Networks Learn. Syst.}, 
  title={Dual-Channel Adaptive Scale Hypergraph Encoders With Cross-View Contrastive Learning for Knowledge Tracing}, 
  year={2025},
  volume={36},
  number={4},
  pages={6752-6766}}

@article{FanWZ2024,
  author={Fang, Yueting and Wu, Hao and Zhao, Yiji and Zhang, Lei and Qin, Shaowei and Wang, Xin},
  journal={IEEE Trans. Neural Networks Learn. Syst.}, 
  title={Diversifying Collaborative Filtering via Graph Spreading Network and Selective Sampling}, 
  year={2024},
  volume={35},
  number={10},
  pages={13860-13873}}

@String{Computing = "Computing" }

@String{Computer = "{IEEE} Computer" }

@ArtifactSoftware{R,
    title = {R: A Language and Environment for Statistical Computing},
    author = {{R Core Team}},
    organization = {R Foundation for Statistical Computing},
    address = {Vienna, Austria},
    year = {2019},
    url = {https://www.R-project.org/},
}

\clearpage

\end{document}